\documentclass{article}

\usepackage{arxiv}

\usepackage[utf8]{inputenc} 
\usepackage[T1]{fontenc}    
\usepackage{hyperref}       
\usepackage{url}            
\usepackage{booktabs}       
\usepackage{amsfonts}       
\usepackage{nicefrac}       
\usepackage{microtype}      
\usepackage{lipsum}		
\usepackage{graphicx}
\usepackage[style=numeric,sorting=none]{biblatex}
\usepackage{doi}
\usepackage{lineno}
\usepackage{url}
\usepackage{pdflscape}
\usepackage{adjustbox}
\usepackage{caption}
\usepackage{subcaption}
\usepackage{longtable}
\usepackage{enumitem}
\usepackage{xcolor}

\title{EEG-ConvTransformer for Single-Trial EEG based Visual Stimuli Classification}

\author{{Subhranil Bagchi}\\
	Department of Computer Science and Engineering\\
	Indian Institute of Technology Ropar\\
	Ropar, India\\
	\texttt{2018csy0002@iitrpr.ac.in} \\
	\And
	{Deepti R. Bathula} \\
	Department of Computer Science and Engineering\\
	Indian Institute of Technology Ropar\\
	Ropar, India\\
	\texttt{bathula@iitrpr.ac.in} \\
}

\date{}

\renewcommand{\headeright}{Preprint}
\renewcommand{\undertitle}{}
\renewcommand{\shorttitle}{EEG-ConvTransformer for Single-Trial EEG based Visual Stimuli Classification}

\hypersetup{
pdftitle={EEG-ConvTransformer for Single-Trial EEG based Visual Stimuli Classification},
pdfsubject={cs.CV, cs.LG, eess.SP},
pdfauthor={Subhranil Bagchi, Deepti R. Bathula},
pdfkeywords={EEG, Visual stimuli classification, Deep learning, Transformer, Multi-head attention, Inter-region similarity, Temporal convolution,  Inter-head diversity, Head representations},
}

\begin{document}
\maketitle

\begin{abstract}
	Different categories of visual stimuli activate different responses in the human brain. These signals can be captured with EEG for utilization in applications such as Brain-Computer Interface (BCI). However, accurate classification of single-trial data is challenging due to low signal-to-noise ratio of EEG. This work introduces an EEG-ConvTranformer network that is based on multi-headed self-attention. Unlike other transformers, the model incorporates self-attention to capture inter-region interactions. It further extends to adjunct convolutional filters with multi-head attention as a single module to learn temporal patterns. Experimental results demonstrate that EEG-ConvTransformer achieves improved classification accuracy over the state-of-the-art techniques across five different visual stimuli classification tasks. Finally, quantitative analysis of inter-head diversity also shows low similarity in representational subspaces, emphasizing the implicit diversity of multi-head attention.
\end{abstract}


\section{Introduction}
\label{section:intro}
Different categories of visual stimuli evoke a significantly distinct pattern of responses in the human brain. An overview of the earliest literature \cite{grill2003neural} indicate that faces of humans and animals activate similar region called the \textit{Fusiform Face Area} (FFA) \cite{kanwisher1999fusiform} compared to scenes, which activate \textit{Parahippocampal Place Area} (PPA) \cite{epstein1998cortical}, and various man-made objects. In addition to these maximally responsive regions, the non-maximally responsive regions may also produce category-distinct patterns. This indicates that category-wise representations are not restricted to particular regions but are widely distributed over the ventral temporal cortex and are overlapping in nature, forming a topographical organization \cite{haxby2001distributed}. These representational patterns across different categories of visual stimuli can be associated with the \textit{Representational Dissimilarity Matrices} (RDMs) \cite{kriegeskorte2008representational}. The RDMs can be utilized to quantitatively compare the different brain regions for the activation patterns they generate as responses to different stimuli. A robust visual stimuli classification system could exploit a similar technique.

Although most of the earlier literature employed functional Magnetic Resonance Imaging (fMRI) to procure data, Electroencephalography (EEG), being a cheap, simplistic and non-invasive technique, presents itself as a befitting alternative. For instance, \cite{kaneshiro2015representational} proposed a confusion matrix based RDM utilizing the EEG data collected from 10 different subjects and large number of trial samples. The work further revealed that electrodes above the right lateral occipital cortex are the most helpful in identifying the individual categories. An efficient EEG based visual stimuli identification system could, thus, not only provide an insight regarding discrimination at the categorical or exemplar level but also be utilized for applications of image labelling and retrieval \cite{parekh2017eeg,wang2009brain}. However, the EEG data render poor spatial resolution \cite{srinivasan1999methods}, as the signals generated at the neuronal level travel through a few resistive mediums to reach the electrodes, a process known as \textit{volume conduction} \cite{jackson2014neurophysiological}. As a result, the recorded signals at the individual electrodes are already a smoothened mixture of multiple signals coming from different underlying sources. This mixing and smoothing makes the underlying spatial structure extremely hard to comprehend and thwarts the classification performance.

The classification of EEG signals, in general, has traditionally relied upon manual feature extractors and classifiers \cite{kaneshiro2015representational,karimi2020temporal,herman2008comparative}, but none of these feature extractors perform near optimally. The deep learning architectures, although newly introduced, have superseded their performances \cite{gao2021attention,schirrmeister2017deep,spampinato2017deep}. Convolutional Neural Networks (CNNs), Deep Belief Networks (DBNs), and Recurrent Neural Networks (RNNs) are the primarily used architectures \cite{roy2019deep,craik2019deep}, with sparing, but increasing involvement of attention-based models \cite{kalafatovich2020decoding,cisotto2020comparison}. The \textit{Transformer Network} \cite{vaswani2017attention} that employs \textit{Multi-Head Attention (MHA)}, has recently been introduced to EEG systems \cite{qu2020residual}. While most of the works on transformers reverberated the temporal domain, employing the multi-head attention over the spatial ordinate to encompass the inter-region representational similarities may be more beneficial. This idea is further endorsed by literature on other EEG signal classification systems, where the brain connectivity and inter-channel relationships are hinted to be beneficial \cite{moon2018convolutional,cisotto2020comparison}.

Inspired by these findings, our work attempts to leverage both spatial and temporal context in terms of inter-region interactions and intra-region temporal patterns, respectively. The main contributions of this work are as follows: \textbf{(1)} We propose a novel \textbf{EEG-ConvTransformer Network}, a deep learning network that employs a series of `ConvTransformer' modules consisting of \textit{multi-head attention} and \textit{convolutional feature expansion} to learn the inter-region representational similarities along with intra-region temporal activities, \textbf{(2)} Present three variants of the proposed network for different parameters, \textbf{(3)} Demonstrate the efficacy of the proposed approach with consistent improvement over state-of-the-art techniques for both category and exemplar level classifications using single-trial EEG signals, and \textbf{(4)} Analyse the head representations to measure the inter-head diversity for the proposed variants.

\section{Related work}
\label{section:litreview}

We segregate the related literature into two sections. The section~\ref{subsection:oldmethods} reviews the existing literature on EEG based classification systems for visual stimuli identification, whereas the section~\ref{subsection:transformer} discusses the developments in the transformer architecture for different applications.

\subsection{Visual stimuli classification using EEG}
\label{subsection:oldmethods}

Majority of the traditional EEG based classification systems rely on two separate steps, namely, feature selection and classifier training, rather than an ``end-to-end'' trainable model; visual stimuli classification is no exception. For instance, \cite{kaneshiro2015representational} proposed Principal Component Analysis (PCA) to extract feature vectors of the specific size from the minimally preprocessed EEG signals and then trained a Linear Discriminant Analysis (LDA) based classifier. \cite{karimi2020temporal} explored the discriminative capabilities of numerous statistical and mathematical features. The experiments conducted on three datasets revealed that multi-valued features such as Wavelet coefficients and the frequency band $\theta$ perform better. One commonly used feature extractor is the Event-Related Potential (ERP), and its general pipeline includes the extraction of the ERP components, namely, \textbf{P1}, \textbf{N1}, \textbf{P2a} and \textbf{P2b}, followed by a classifier such as LDA or Support Vector Machine (SVM) \cite{qin2016classifying,bobe2018single}. Most works use ERP on single-trial classification, but that yields a rather poor performance due to the low \textit{signal-to-noise} ratio of the EEG data. So, ERP components extraction have been extended to multi-trial based classification, using the average superposition of multiple trials \cite{zheng2020evoked}.

Compared to traditional methods, deep learning eliminates the need for manual feature extraction, although manual features could still be passed as inputs \cite{parekh2017eeg}. For most EEG applications, it is observed that shallow models yield good results, and deeper models may lead to performance loss \cite{schirrmeister2017deep,roy2019deep,craik2019deep}. Particularly to EEG based object classification, although shallow architectures of CNNs, with few parameters, have been proposed generously: ConvNet, EEGNet \cite{lawhern2018eegnet} and other variants \cite{kalafatovich2020decoding}, one recent study by \cite{bagchi2021adequately} exemplifies the increased performance for increased width and inclusion of residual connections \cite{he2016deep}. \cite{jiao2019decoding} transforms the EEG signals into images of spectral maps and applies AlexNet-like architectures. Apart from the CNNs, different RNN architectures have also been proposed. Initially proposed by \cite{spampinato2017deep}, these RNN models constitute Long Short-Term Memory (LSTM) modules, and often attention modules \cite{zheng2021attention}. Although few of these techniques applied spatial attention for particular regions, none explored the inter-region representational similarities.

\subsection{Transformer Network}
\label{subsection:transformer}

The vanilla Transformer Network, proposed by \cite{vaswani2017attention}, was a significant development in Natural Language Processing. It introduced a self-attention based fully-attentional model by commissioning the \textit{Multi-Head Attention} for language translation tasks. Inside the multi-head attention, each self-attention head calculates the dot-product based similarities for two tensors (called queries and keys) through matrix multiplication, scaling and Softmax. Originally proposed to eradicate the use of RNNs and invoke parallelization, the idea has been extended well to substitute CNNs in Computer Vision. The Vision Transformer relies on cropping the original image into multiple patches, which pass through the stack of transformer layers \cite{dosovitskiy2020image}. Unlike CNNs, this eliminates the concept of ``receptive field'', instead focuses on the patch to patch interaction at a global scale. Additionally, a very recent work introduces ConvTransformer that proposes the convolutional multi-head attention for image frame synthesis \cite{liu2020convtransformer}.

Unlike word vectors or image patches, EEG data consist of temporal samples from multiple channels. One way to apply transformers would be to treat each time samples' pairs as individual ``query-key'' pairs. These architectures, much similar to a vanilla transformer, can be applied directly as Encoder-Decoder models for EEG systems \cite{qu2020residual,krishna2019eeg}. However, if one were to consider transformer architecture to classify visual stimuli, the spatial relations become much more meaningful. Only one prototypical work, recently proposed for motor imagery classification, employs multi-head attention between electrode pairs to extract the spatial similarities \cite{sun2021eeg}. To the best of our knowledge, transformer networks, exploring the region-to-region interactions, have not been explored in EEG based visual stimuli classification system.

\section{Proposed methodology}
\label{section:method}

This section describes the modules of the proposed EEG-ConvTransformer Network. We start with the conversion of spatio-temporal EEG signal into time-frames of activity-maps in section~\ref{subsection:aep}. Next, the need for localized feature extraction for learning the lower-level representations is explained in section~\ref{subsection:localcnn}. Following on, the proposed ConvTransformer module to assess the higher-level, inter-region representational similarities is explained in section~\ref{subsection:convtransformer}. Finally, in section~\ref{subsection:together}, the assembly of individual modules that conforms to the overall architecture is described.

\subsection{Azimuthal Equidistant Projection}
\label{subsection:aep}

\begin{figure*}[t]
\centerline{\includegraphics[width=0.75\linewidth]{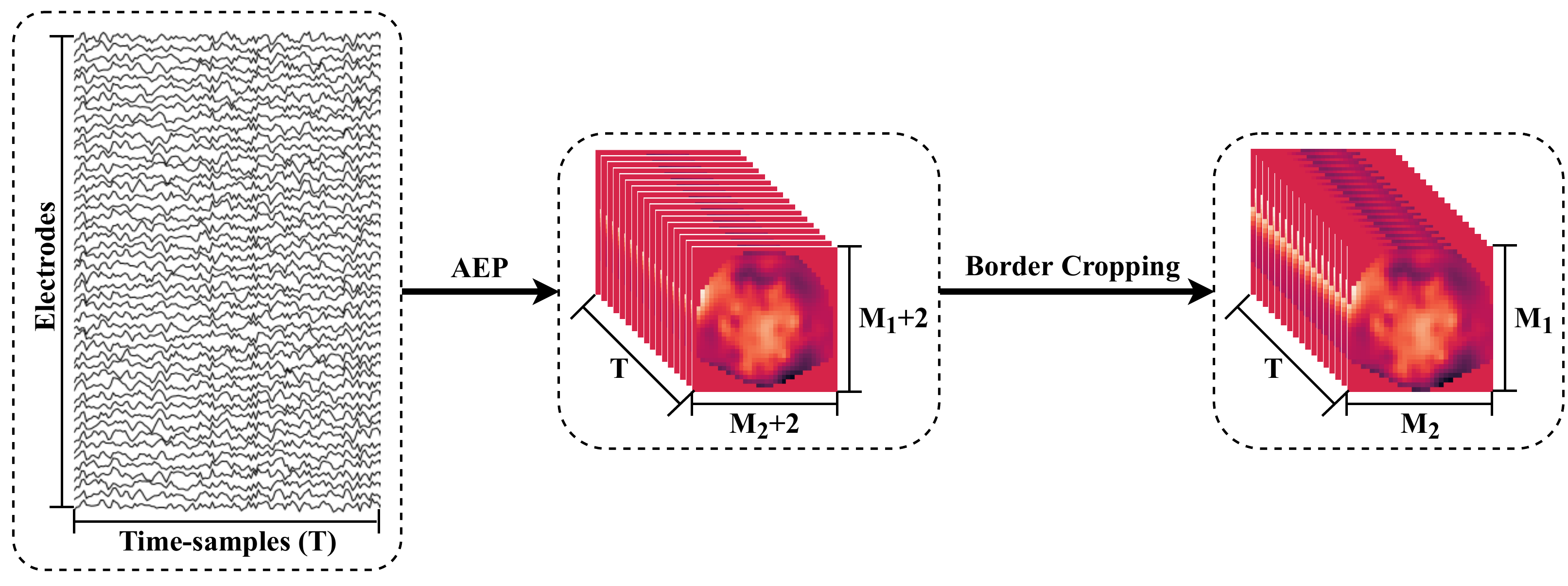}}
\caption{Formation of border cropped multi-frame activity-maps from the AEP of the preprocessed EEG signal.}
\label{fig:1}
\end{figure*}

To transform the multi-channel EEG signals to time-frames $(T)$ of image-like activity maps, the idea of Azimuthal Equidistant Projection (AEP) \cite{snyder1987map}, as proposed by \cite{bashivan2015learning} for EEG based load classification, is utilized. The AEP allows the relative distances between the neighboring electrodes to be preserved while projecting them from their location in 3-dimensional space to a 2-dimensional surface. Following their work, the in-between electrode values on the plain are interpolated using the  Clough-Tocher scheme \cite{alfeld1984trivariate}. However, contrary to the three frequency power bands from the earlier work, the AEP and interpolation are applied to the preprocessed signal to form a single channel mesh of $G_1\times G_2$ per time-frame. For simplicity, we considered $G_1=G_2$. Further, border cropping is applied, resulting the mesh to be of size $(G_1-2)\times(G_2-2)$ or $M_1\times M_2$ per time-frame, such that $G_1=M_1+2$, $G_2=M_2+2$ and $M_1=M_2$, as shown in Figure ~\ref{fig:1}.

\subsection{The Local Feature Extractor module}
\label{subsection:localcnn}

After projecting the EEG signals to image-like time-frames, the already spatially smoothened data is smoothened further due to the interpolation. As a result, the first objective is to eliminate the unnecessary computational costs for these smoothened images and extract the lower level temporal features from every vicinity of electrodes, i.e., local feature extraction. This objective is achieved through a 3-dimensional convolutional (Conv) layer with appropriate spatial stride.

\begin{figure*}[t]
\centerline{\includegraphics[width=.75\linewidth]{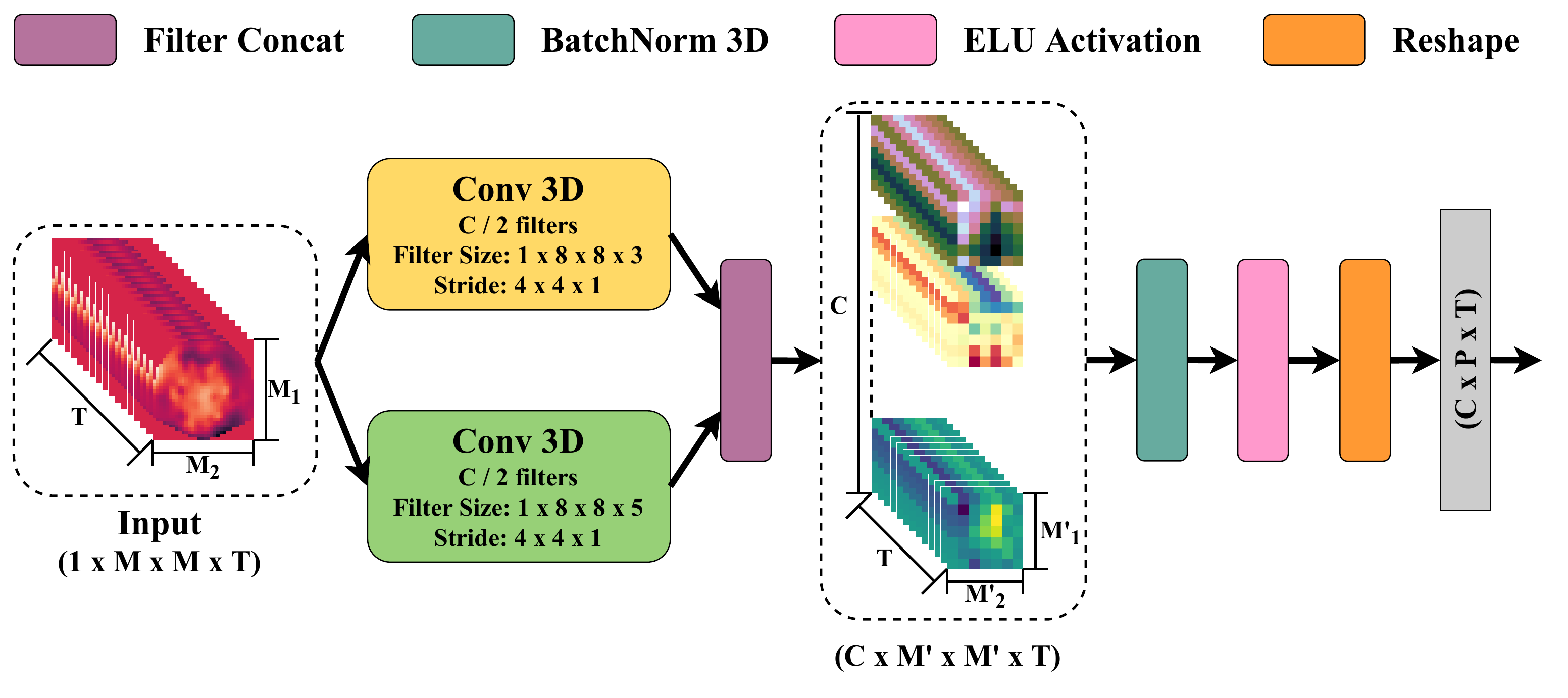}}
\caption{Visualization of the LFE module.}
\label{fig:2}
\end{figure*}

For an input sample, $x\in \mathcal{R}^{1\times M_1\times M_2\times T}$, $M_1$, $M_2$ and $T$ are the mesh height, mesh width and time-frames, respectively. Let us consider a point $(1,m,m',t)$ on $x$, such that $m\in [1, s+1, 2s+1,..., M_1-2s+1]$, $m'\in [1, s+1, 2s+1,..., M_2-2s+1]$, for spatial stride $s$, and $t\in [1, 2, 3,..., T]$, and the kernel $k^c\in \mathcal{R}^{1\times k_m\times k_m'\times k_t}$, such that $k_m=k'_m=2s$ and $c\in [1, 2, 3,..., C]$ represents the specific output kernel.

Then the convolution operation on $x_{(1,m,m',t)}$ by $k^c$ is formulated as\textemdash
\begin{equation}
Conv(x,k^c)_{1,m,m',t} = \sum^{k_m}_{h=1} \sum^{k_m'}_{w=1} \sum^{k_t}_{r=1} x_{m+h-1,m'+w-1,t+r-1}k^{c}_{h,w,r} + b^c
\label{equation:conv}
\end{equation}
where $b^c$ represents the bias associated with the kernel. The stride $s$ introduces the convolution operations for overlapping spatial patches. The number of kernels is chosen to be $C/2$ for each value of $k_t\in [3, 5]$, followed by the output channel concatenation. Padding retains the original temporal dimension. So the output results in $x^{conv}\in \mathcal{R}^{C\times M'_1\times M'_2\times T}$, for $M'_1\ll M_1$ and $M'_2\ll M_2$. The 3-dimensional Conv layer, followed by Batch Normalization layer (BatchNorm) \cite{ioffe2015batch} and Exponential Linear Unit (ELU) activation, together forms the Local Feature Extractor (LFE) module $M^{[LFE]}$, and outputs $x^{[LFE]}$. For a batch of size $B$, a new major axis is augmented to all the input-output operations to reference every sample. The data flow for this module is depicted in Figure~\ref{fig:2}. As visualized, these extracted local features, for each sample $x^{[LFE]}$, gets reshaped as $\mathcal{R}^{C\times (M'_1.M'_2)\times T}$ or $\mathcal{R}^{C\times P\times T}$, with $P$ being the number of patches, and this reshaped output gets forwarded to the ConvTransformer module.

\subsection{The ConvTransformer module}
\label{subsection:convtransformer}

Unlike the previous transformers \cite{vaswani2017attention,dosovitskiy2020image} that prefer working on either temporal or spatial axis, the proposed ConvTransformer (CT) module attempts to extract both the spatial relationships and temporal features. Consequently, the Multi-Head Attention sub-module that learns the inter-region representational similarities through a number of self-attention heads is succeeded by a Convolutional Feature Expansion (CFE) sub-module that extracts meaningful temporal information. Both of these sub-modules are described in this section. The attentions, here, being applied to regions instead of electrodes, makes it different from the one proposed by \cite{sun2021eeg}. This step significantly reduces the computational complexity for data from high-density electrode devices and is practical due to the volume conduction involved with the EEG data.

\subsubsection{Multi-Head Attention}
\label{subsubsection:mha}

\begin{figure*}[t]
\centerline{\includegraphics[width=0.75\linewidth]{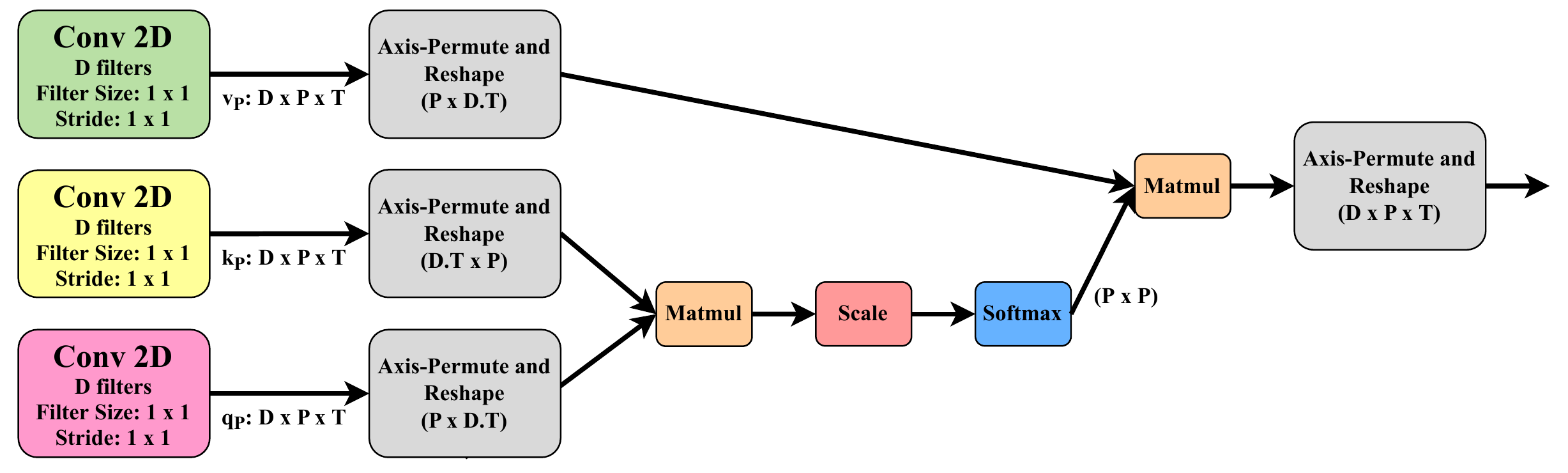}}
\caption{The operations inside a single self-attention head.}
\label{fig:3}
\end{figure*}

\begin{figure*}[t]
\centerline{\includegraphics[width=.95\linewidth]{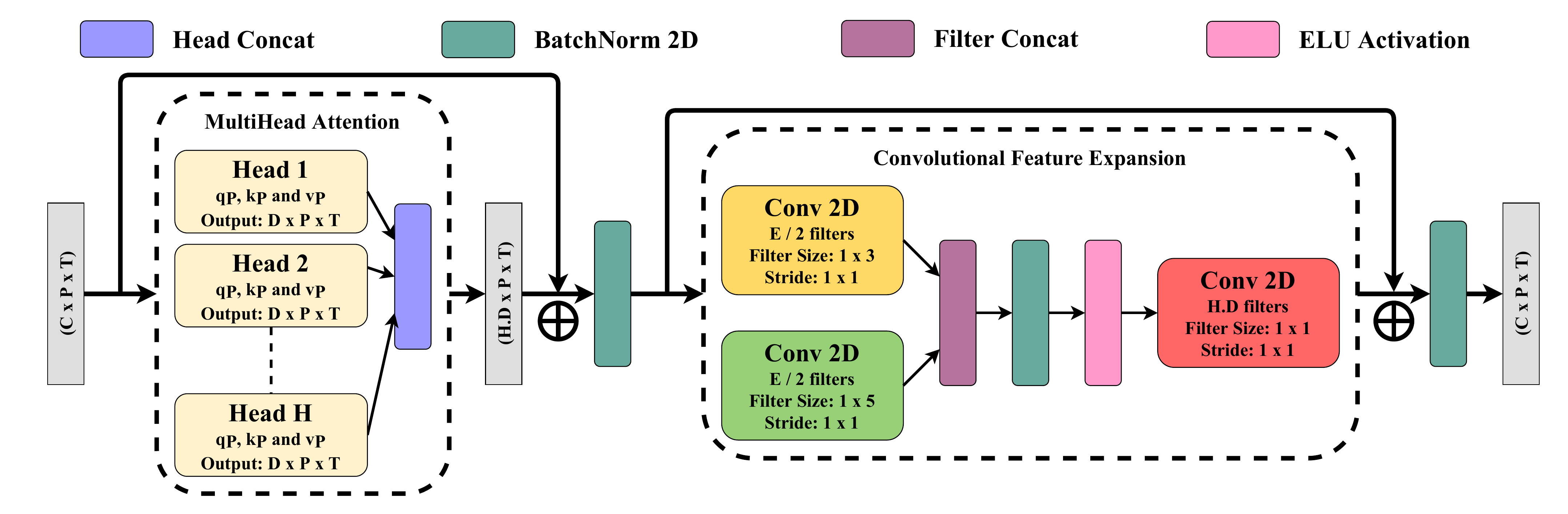}}
\caption{Visualization of the proposed EEG-ConvTransformer module in which the MHA module is succeeded by a CFE module.}
\label{fig:4}
\end{figure*}

Each self-attention head, $h\in[1, 2,..., H]$, with $H$ being the total number of heads, relies on $q_P$ (query), $k_P$ (key) and $v_P$ (value) vectors for patch assessment. The patch representations are projected to latent representations of $q_P, k_P, v_P\in\mathcal{R}^{D\times P\times T}$, through point-wise convolution operations \cite{Chollet_2017_CVPR}, such that $D \ll  C$. The point-wise convolution operation refers to convolution with a kernel size of $1$, which, without any bias, encodes individual patches to smaller channel spaces. This encoding reduces the computational complexity of the matrix-multiplications invoked right after these latent representations permute axis-wise and reshape such that $q_P, k_P, v_P\in\mathcal{R}^{P\times D.T}$. The ``query-key'' pair aims to map the key patches to the query patches based on their in-between representational similarity, which is calculated as their scaled dot-product, followed by Softmax operation \cite{vaswani2017attention}. The resultant matrix is again multiplied with $v_P$ to calculate the \textit{representational context} as the aggregation of the self-attentional interactions. For a particular patch $P$ and a single self-attention head $h$, the representational context is calculated as\textemdash

\begin{equation}
Head(q_P,k_P,v_P)^h_x = Softmax(\frac{q_P.k'_P}{\sqrt{D.T}}).v_P
\label{equation:head}
\end{equation}
where $k'_P$ is the transpose of $k_P$. Finally, this output is axis-wise permuted and reshaped as $\mathcal{R}^{D\times P\times T}$. These operations are graphically depicted in Figure~\ref{fig:3} for better understanding. Multiple projections of $q_P, k_P$ and $v_P$ calculate the respective self-attention heads and their outputs get concatenated along the channel axis to form the aggregate of multiple heads, as given by\textemdash
\begin{equation}
MultiHead(q_P,k_P,v_P)_x = Concat(Head_1, Head_2,..., Head_H)
\label{equation:mha}
\end{equation}
The $D$ is calculated as $C/H$, so that, post concatenation, the output retains the initial size of $\mathcal{R}^{C\times P\times T}$. As the output of the MultiHead module, $x^{[mh]}$, is combined with its input as part of residual mapping, followed by BatchNorm to obtain the sub-module output, such that $x^{[MHA]} = BN(x^{[mh]} + x^{[in]})$, retaining the original dimension is essential \cite{he2016deep}. The residual mapping is easier to optimize than an unreferenced one and allows better gradient flow throughout the network. Unlike the other transformers, we utilized BatchNorm, instead of LayerNorm, following the trend of CNNs for EEG systems \cite{lawhern2018eegnet,kalafatovich2020decoding}. The complete MHA sub-module is visualized in Figure~\ref{fig:4}.

\subsubsection{Convolutional Feature Expansion (CFE)}
\label{subsubsection:convexp}

The output of the MHA module, $x^{[MHA]}\in\mathcal{R}^{C\times P\times T}$, gets forwarded to a temporal convolutional sub-module to extract patch-wise temporal features. The convolution operations performed by every kernel $k^e\in \mathcal{R}^{C\times 1\times k_t}$, such that $e\in[1, 2,..., E]$ for the expanded number of channels $E$, should extract sufficient temporal features to aid the classification. Like the LFE module, for each value of $k_t\in [3, 5]$, $E/2$ number of channels are generated, followed by channel concatenation. This Conv layer is followed by BatchNorm and ELU non-linearity. Next, a point-wise convolution maps the temporally filtered features to the initial channel space $C$, as required for residual mapping, followed by a BatchNorm. Together, these layers form the Convolutional Feature Expansion (CFE) sub-module, visualized in Figure~\ref{fig:4}.

\subsection{Architectural design}
\label{subsection:together}

\begin{figure*}[t]
\centerline{\includegraphics[width=.75\linewidth]{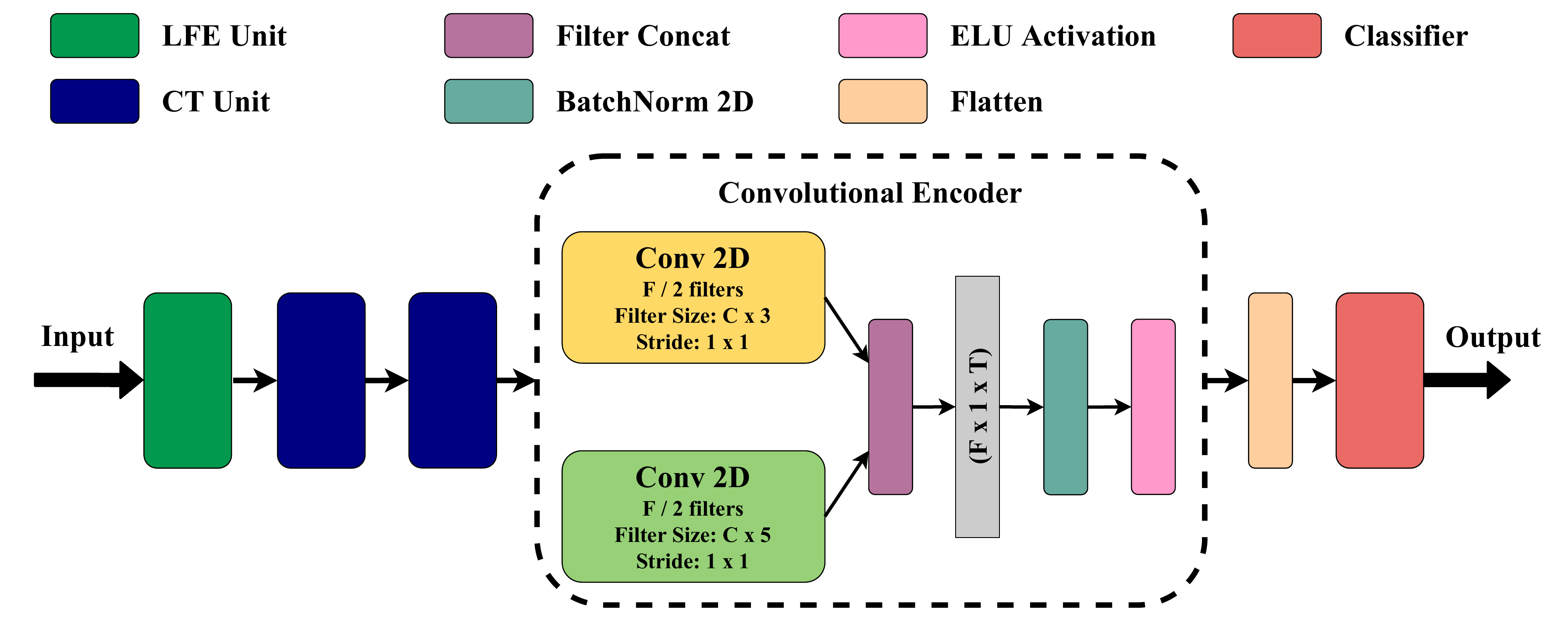}}
\caption{The generalized architecture of the proposed network.}
\label{fig:5}
\end{figure*}

After the first module, i.e., the LFE, the latent representation must pass through a series of CT modules. As established in the literature, shallow networks perform better \cite{roy2019deep,schirrmeister2017deep}, so a series of only two of such modules is stacked. The transformer models generally use positional encoding to represent the location or position of the token or patch in the input sequence as the tasks depends on them. However, such encoding is irrelevant for EEG-ConvTransformer as the patches should remain fixed in their respective locations on the scalp for the entire trial set. Post the series of CT modules, a simple \textit{Convolutional Encoder} module, consisting of a 2-dimensional Conv layer, a BatchNorm layer and ELU activation, temporally convolves and encodes the channels of all the patches to a new channel space $F$, through kernels $k^{f}\in \mathcal{R}^{C\times P\times k_t}$, for all $f\in[1, 2,..., F]$. Like all previous convolutions, each value of $k_t\in [3, 5]$ generates $F/2$ number of channels, which are then concatenated. This module, as part of the final network architecture, is visualized in Figure ~\ref{fig:5}. To make the final prediction, the classifier, chosen identical to that of the 1-D Wide-ResNet \cite{bagchi2021adequately}, has a series of three fully-connected layers with output sizes of 500, 100 and number of classes, respectively. Each intermediate layer of the classifier is followed by Dropout with a probability of $0.5$ \cite{srivastava2014dropout} and Rectified Linear Unit (ReLU) activation.

\section{Experiments and results}
\label{section:experiment}

In the section~\ref{subsection:eegdataset} we describe the dataset used for experimentations. Following that, in subsequent sections~\ref{subsection:architecture} and~\ref{subsection:training}, we explain the parameter choices of the proposed variants and the training strategies, respectively. Finally, in section~\ref{subsection:eval} we then evaluate the performance of the proposed variants, followed by an in-depth analysis of the inter-head diversity in section~\ref{subsection:analysis}.

\subsection{Visual stimuli EEG dataset}
\label{subsection:eegdataset}

An effective deep learning architecture relies heavily on the availability of a considerable amount of data. Towards this end, we utilized the public EEG data from the Stanford Digital Repository \cite{kaneshiro2015dataset}.

As described by \cite{kaneshiro2015representational}, this dataset consists of samples from $10$ subjects with normal color vision, who were shown color photographs (with mid-grey background) of $72$ exemplar stimuli for six different categories, with each category comprising $12$ unique exemplars. The categories are, namely, Human Body (HB), Human Face (HF), Animal Body (AB), Animal Face (AF), Fruit-Vegetable (FV), and Inanimate Object (IO). For the data collection, a dark and acoustically shielded booth was used. The participants sat on a chair that was $57$
cm from the desk on which the monitor flashed the images of stimuli in random order for a period of $500$ ms of each trial. An inter-trial gap of $750$ ms was also introduced. The data was recorded as blocks of trials, with each block consisting of $864$ trials ($12$ trials per-stimuli), with gaps provided in-between every $36$ trials. The per-subject data was recorded over two experimental sessions of around a week apart, with each containing three blocks. This paradigm resulted in $72$ trials per stimuli and around 5,184 trials per subject.

The EEG data procured through an unshielded $128$ channel EGI HydroCel Geodesic Sensor Net \cite{tucker1993spatial} was sampled at a frequency of $1$ kHz. As part of preprocessing, high-pass fourth-order Butterworth filter and low-pass eighth-order Chebyshev Type I were used to remove frequency components below $1$ Hz and above $25$ Hz, respectively. Further, the filtered data was down-sampled at $62.5$ Hz, and channels $125-128$ were removed. The data was average referenced after per-subject ocular artifacts were removed using extended \textit{Infomax ICA}. The clean data was finally partitioned into trials of $32$ time-samples (post-stimulus response) \cite{kaneshiro2015representational}.

\subsection{Architectural parameters}
\label{subsection:architecture}

For the purpose of implementation, the initial mesh size $G_1\times G_2$ was considered to be $34\times34$, such that after border cropping, the mesh results in size $32\times32$. The spatial kernel sizes $k_m$ and $k_m'$ for the LFE module $M^{[LFE]}$ were taken to be $8$, so the stride $s$ becomes $4$. This choice of parameters ensures that the mesh of $32\times32$ shrinks down to a much smaller $7\times7$ (denoted as $M'_1\times M'_2$). As a result, the output from this module consists of $P=49$ number of representations.

\begin{table}[t]
\caption{Model Parameters with respect to each variant}
\begin{center}
\begin{tabular}{p{0.25\textwidth}p{0.1\textwidth}p{0.1\textwidth}p{0.1\textwidth}}
\hline
\textbf{Parameter} & \textbf{CT-Slim} & \textbf{CT-Fit} & \textbf{CT-Wide}\\
\hline
Heads $(H)$ & $4$ & $8$ & $12$\\
Projections $(D=H/2)$ & $2$ & $4$ & $8$\\
Local Features $(C=H.D)$ & $8$ & $32$ & $72$\\
Expansions $E=(C.H/2)$ & $16$ & $128$ & $432$\\
Final Channels $(F=64.H)$ & $256$ & $512$ & $768$\\
Parameter Size* & $4.56$M & $11.52$M & $23.55$M\\
\hline
\multicolumn{4}{l}{* \textit{with respect to 72 class task.}}
\end{tabular}
\label{tab_param}
\end{center}
\end{table}

Three variants of the EEG-ConvTransformer network are proposed for experimentation, varying in the number of heads and width of the network. These are, namely, CT-Slim, CT-Fit and CT-Wide. Table~\ref{tab_param} showcases the different parameters involved with the variants: the number of local features $C$, projected channels $D$, expanded channels $E$, and final channels $F$, all calculated in terms of the number of heads, as well as the parameter sizes. Earlier works suggest that thicker networks perform better for EEG based visual stimuli detection, but come at the cost of additional parameters and increased computation \cite{bagchi2021adequately}. The choice of parameters is directly related to the width of the network, increasing from slim to fit to wide. 

\subsection{Training protocols}
\label{subsection:training}

Our models were trained using Adam optimizer \cite{DBLP:journals/corr/KingmaB14}, a variant of Stochastic Gradient Descent (SGD) that relies on the first and second moments of the gradient to achieve significantly faster convergence, along with weight-decay regularization. The cross-entropy loss was minimized, and to help with the convergence, the learning rate was multiplied by a decay factor $\gamma$ after every five epochs, starting from the $15^{th}$ epoch, using \textit{Multi-step scheduler}. For information on the training hyper-parameters, refer to section S1 in the supplementary material. Within-subject experiments were conducted using \textit{stratified 10-fold cross-validation}. In this strategy, the per-subject data is split into ten folds so that the per-class samples are uniformly distributed over the folds.

\subsection{Performance evaluation}
\label{subsection:eval}

We evaluated our models for five different classification tasks, namely, 6-class category level classification, 72-class exemplar level classification, 2-class HF vs IO category level classification, 12-class HF exemplar level classification, and 12-class IO exemplar level classification, in order to compare them to the previously proposed methods. Furthermore, as part of the training protocol, the per-task training epochs and weight decays were fine-tuned as per the stratified 10-fold cross-validation.

Table~\ref{tab_result} reports the average subject-wise accuracy and the corresponding sample standard deviation. The CT-Wide variant of the proposed EEG-ConvTransformer exhibits consistent improvement in the classification accuracy over all the previous methods for all the tasks. Interestingly, the improvements observed for increased heads and widths (variants) were considerably better for the exemplar level classifications: 72-Exemplar, HF Exemplar and IO Exemplar, compared to category level classification: 6-Category and HF vs IO Category. 

\begin{table*}[t]
\caption{Comparison of the proposed method with existing literature for different tasks.}
\begin{center}
\begin{tabular}{p{0.22\linewidth}p{0.12\linewidth}p{0.12\linewidth}p{0.12\linewidth}p{0.12\linewidth}p{0.12\linewidth}p{0.12\linewidth}}
\hline
\textbf{Method} & \multicolumn{5}{c}{\textbf{Accuracy (\%)}} \\
\cline{2-6} 
\textbf{} & \textbf{6-Category} & \textbf{72-Exemplar} & \textbf{HF vs IO} & \textbf{HF} & \textbf{IO} \\
\hline
LDA~\cite{kaneshiro2015representational}&                $40.68\pm5.54$&   $14.46\pm6.43$&      $81.06\pm3.66$&    $18.30\pm5.63$&    $28.87\pm10.57$\\
ICA-ERP~\cite{bobe2018single}&                           $43.50$&          \textemdash&         \textemdash&       \textemdash&       \textemdash\\
Shallow \cite{kalafatovich2020decoding}&                 $49.04\pm6.99$&    $23.72\pm10.95$&    \textemdash&       \textemdash&       \textemdash\\
LSTM~\cite{kalafatovich2020decoding,jiao2019decoding}&   $44.77\pm6.30$&    $15.39\pm6.01$&     $80.67$&           \textemdash&       \textemdash\\
LSTM + CNN~\cite{kalafatovich2020decoding}&              $46.18\pm6.79$&    $23.23\pm10.48$&    \textemdash&       \textemdash&       \textemdash\\
CNN~\cite{kalafatovich2020decoding,jiao2019decoding}&    $50.00\pm6.61$&    $25.93\pm10.67$&    $83.10$&           \textemdash&       \textemdash\\
Attention CNN~\cite{kalafatovich2020decoding}&           $50.37\pm6.56$&    $26.75\pm10.38$&    \textemdash&       \textemdash&       \textemdash\\
CNN-ResNet101~\cite{jiao2019decoding}&                   \textemdash&       \textemdash&        $85.50$&           \textemdash&       \textemdash\\
1-D Wide-Res CNN~\cite{bagchi2021adequately}&            $51.29\pm7.57$&    $28.68\pm12.58$&    $88.83\pm3.49$&    $24.64\pm7.90$&    $47.12\pm16.26$\\
\hline
CT-Slim&                                                $51.96\pm8.63$&    $26.08\pm13.68$&    $88.78\pm4.25$&    $25.67\pm8.15$&    $47.53\pm16.28$\\
CT-Fit&                                                 $52.17\pm8.15$&    $27.14\pm13.35$&    $89.58\pm3.97$&    $26.77\pm9.17$&    $49.91\pm17.44$\\
CT-Wide&                                                $\mathbf{52.33\pm8.28}$&    $\mathbf{29.44\pm13.51}$&    $\mathbf{89.64\pm4.16}$&    $\mathbf{27.20\pm9.10}$&    $\mathbf{50.59\pm17.22}$\\
\hline
\multicolumn{6}{l}{\textemdash \ \textit{missing from the work.}}
\end{tabular}
\label{tab_result}
\end{center}
\end{table*}

\subsubsection{Subject level evaluation}
\label{subsubsection:subjecteval}

\begin{figure*}[t]
	\centering
	\begin{subfigure}[h]{.75\textwidth}
		\centering
		\includegraphics[width=\textwidth]{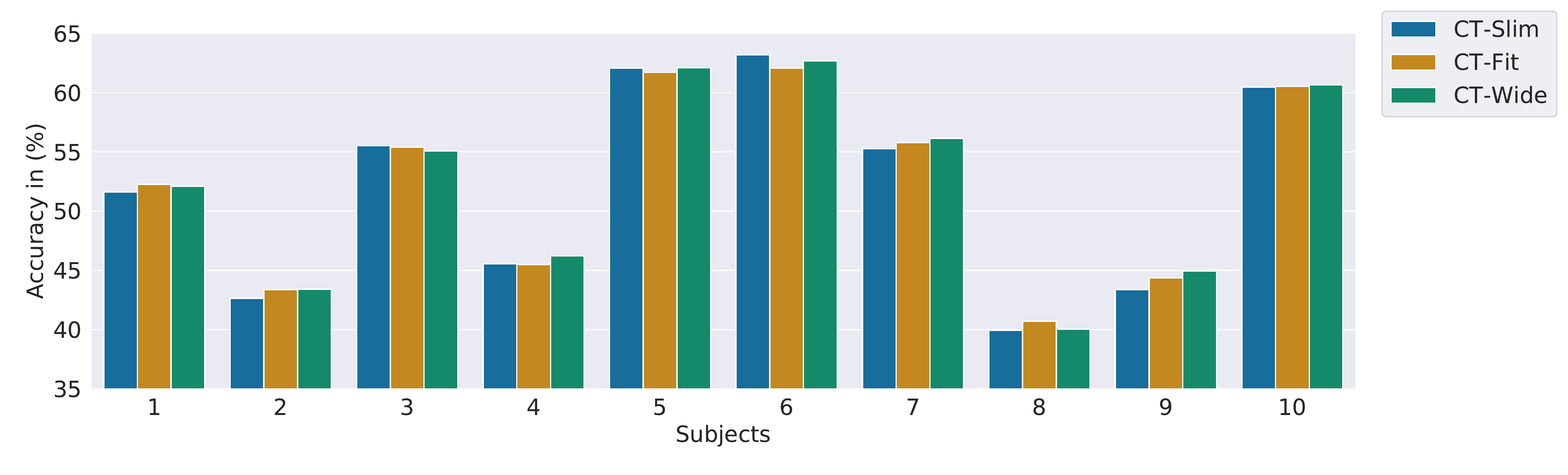}
		\caption{for 6-category classification}
		\label{fig:6a}
	\end{subfigure}
	\begin{subfigure}[h]{.75\textwidth}
		\centering
		\includegraphics[width=\textwidth]{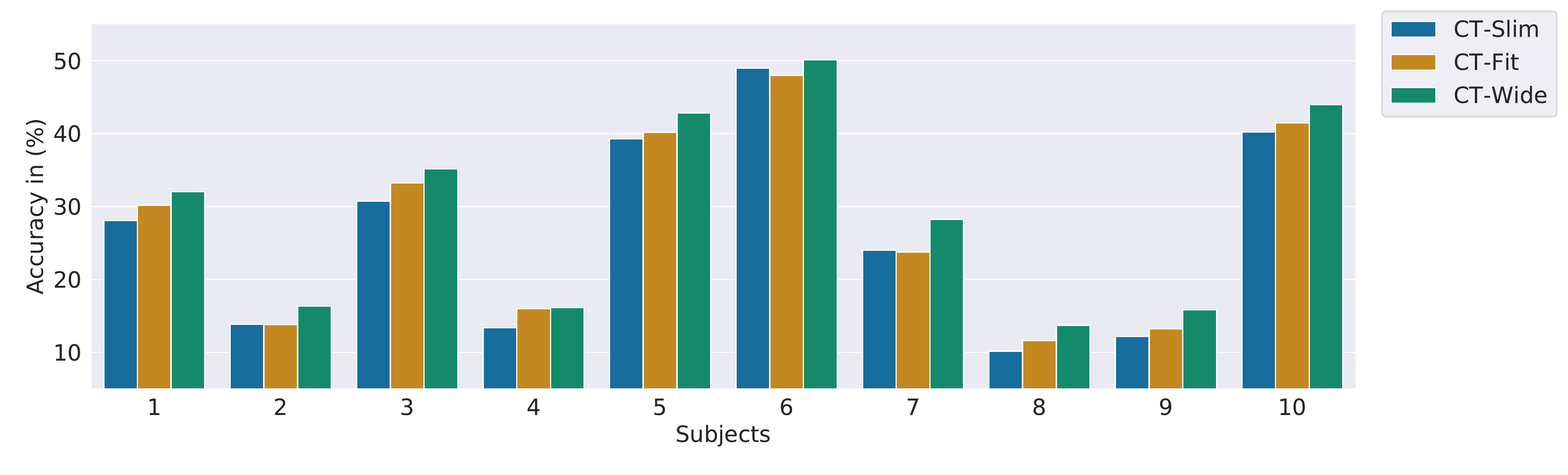}
		\caption{for 72-exemplar classification}
		\label{fig:6b}
	\end{subfigure}
	\caption{Plots for subject wise classification accuracies.}
	\label{fig:6}
\end{figure*}

Figure~\ref{fig:6} depicts the subject wise accuracies attained by the different variants for two of the tasks: 6-category and 72-exemplar classifications. The considerable variations in performances for different subjects could be observed from the plots. For example, the accuracies for subject 8 are consistently the minimum, whereas subject 6 reports the maximum accuracy for four of the tasks, except for HF vs IO, where subject 5 reports the maximum accuracy (remaining plots deferred to section S2 in the supplementary material). Additionally, for 72-Exemplar classification, the extra width of the CT-Wide, benefits the network for all the subjects, in contrast to the 6-category level classification, where the much slimmer network CT-Slim seems to perform on par for subjects 3 and 6.

\subsubsection{Class level evaluation}
\label{subsubsection:classeval}

\begin{figure*}[t]
	\centering
	\begin{subfigure}[h]{0.4\textwidth}
		\centering
		\includegraphics[width=\textwidth]{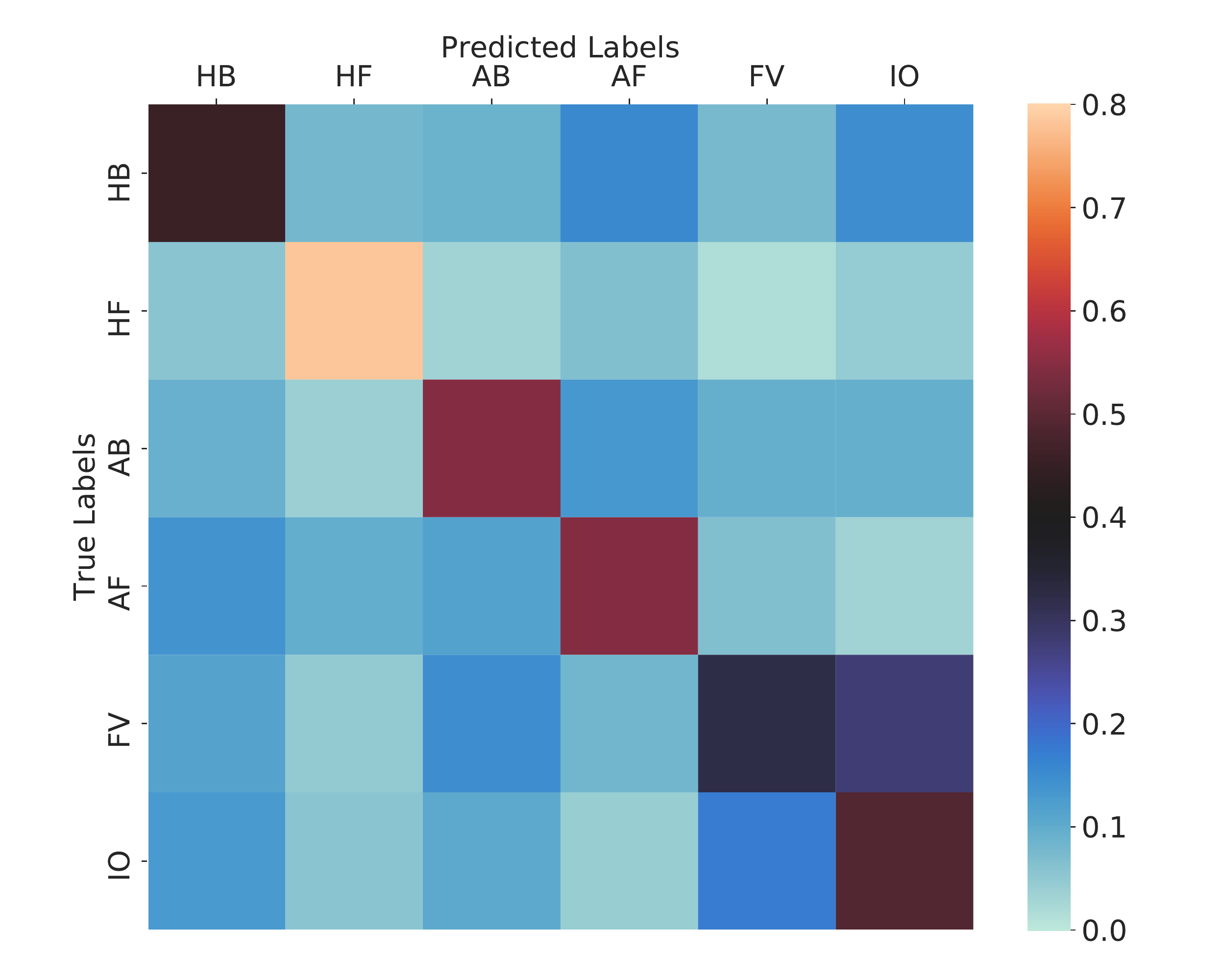}
		\caption{for 6-category classification}
		\label{fig:7a}
	\end{subfigure}
	\begin{subfigure}[h]{0.4\textwidth}
		\centering
		\includegraphics[width=\textwidth]{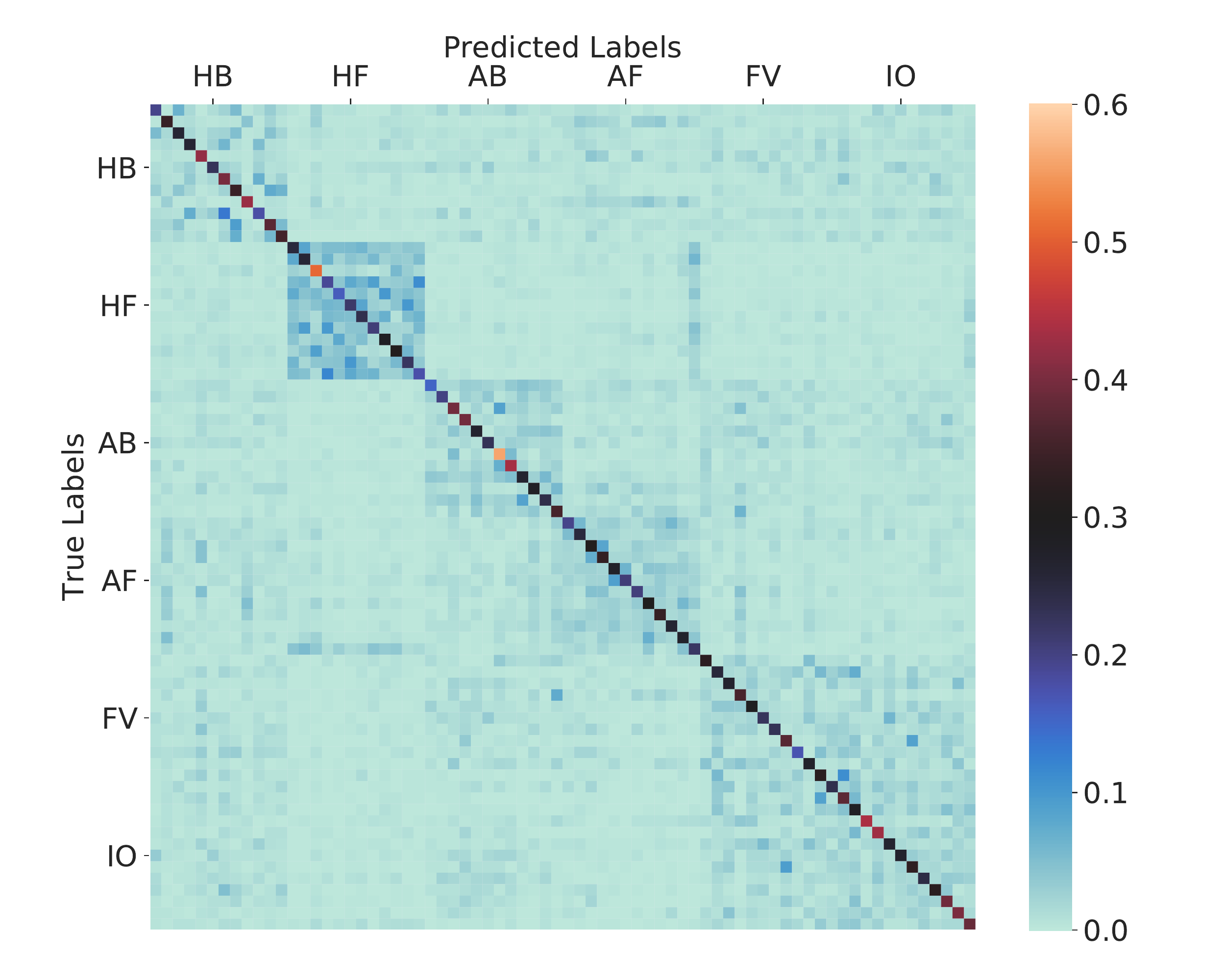}
		\caption{for 72-exemplar classification}
		\label{fig:7b}
	\end{subfigure}
	\caption{Confusion matrices for category and exemplar level classification tasks for CT-Wide.}
	\label{fig:7}
\end{figure*}

To understand the efficacy of our proposed method in predicting the correct class information, we plotted the normalized confusion matrices for 6-category and 72-exemplar classification tasks. The confusion matrices for the proposed CT-Wide variant are given in Figure~\ref{fig:7} (remaining plots deferred to section S3 in the supplementary material). The confusion matrices show consistency with the previous findings by \cite{kaneshiro2015representational}, that the Human Faces, as a category, could be identified most prominently, with a classification accuracy of $78.37\%$ (for CT-Wide), in contrast to the Fruit-Vegetables and Inanimate Objects, the two categories that together form a higher-level ordinate. However, at an exemplar level, the Human Face exemplars have the highest degree of confusion, whereas it was relatively easier to identify the Inanimate Object exemplars, which are also evident from the exemplar level classifications of the respective categories.

\subsection{Inter-head diversity analysis}
\label{subsection:analysis}

\begin{figure*}[t]
	\centering
	\begin{subfigure}[h]{0.75\textwidth}
		\centering
		\includegraphics[width=\textwidth]{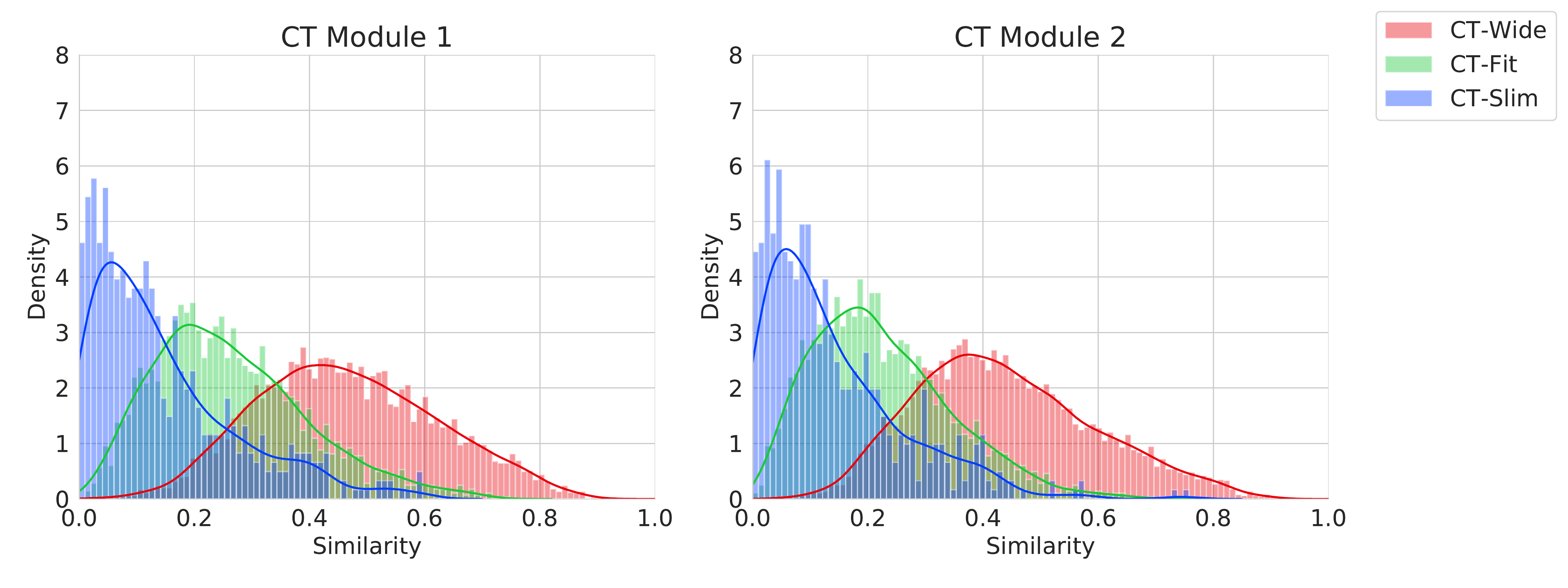}
		\caption{for 6-Category classification}
		\label{fig:8a}
	\end{subfigure}
	\begin{subfigure}[h]{0.75\textwidth}
		\centering
		\includegraphics[width=\textwidth]{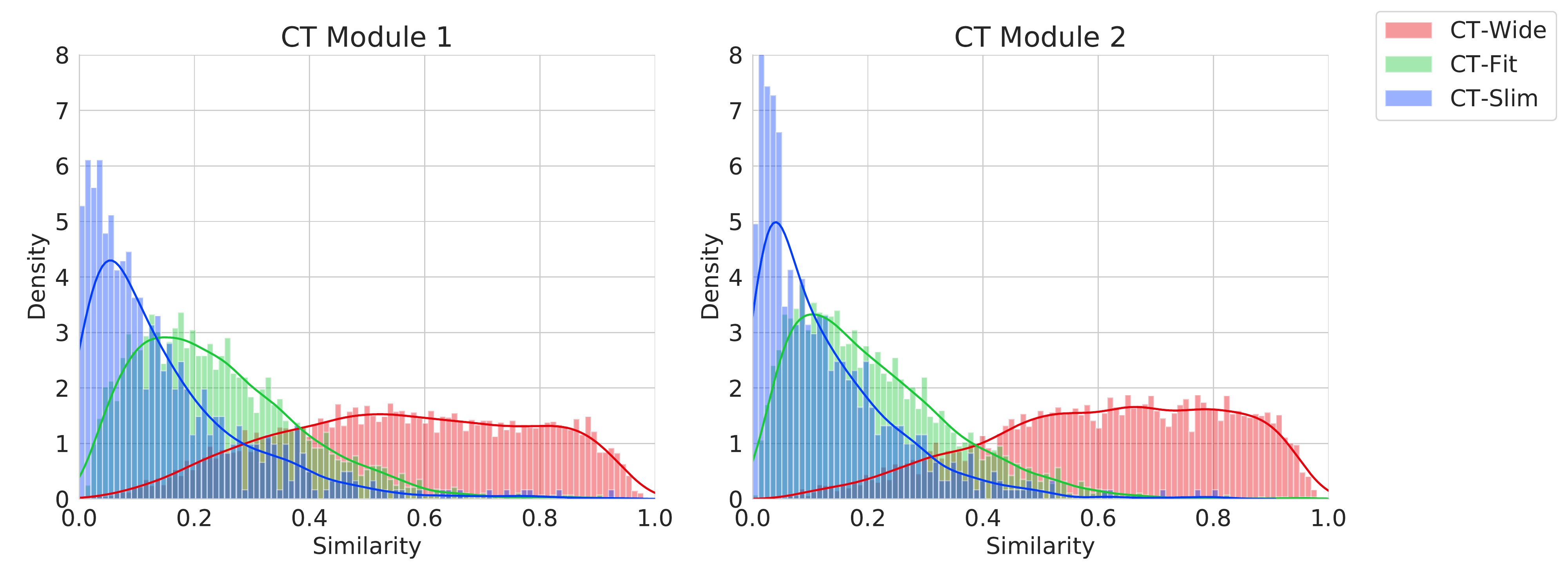}
		\caption{for 72-Exemplar classification}
		\label{fig:8b}
	\end{subfigure}
	\caption{Sample distributions of $CKA_{(HR)}$ using KDEs.}
	\label{fig:8}
\end{figure*}

\begin{table*}[hbt!]
\caption{Mean $CKA_{(HR)}$ values for different tasks and variants.}
\begin{center}
\begin{tabular}{p{0.12\textwidth}p{0.12\textwidth}p{0.07\textwidth}p{0.07\textwidth}}
\hline
\textbf{Task} & \textbf{Variant} & \multicolumn{2}{c}{\textbf{Mean $CKA_{(HR)}$}}\\
\cline{3-4}
\textbf{} & \textbf{} & \textbf{CT 1} & \textbf{CT 2}\\
\hline
                & CT-Slim   & $0.144$   & $0.139$\\
6-Category      & CT-Fit    & $0.263$   & $0.224$\\
                & CT-Wide   & $0.465$   & $0.438$\\
\hline
                & CT-Slim   & $0.145$   & $0.125$\\
72-Exemplar     & CT-Fit    & $0.239$   & $0.209$\\
                & CT-Wide   & $0.565$   & $0.615$\\
\hline
                & CT-Slim   & $0.149$   & $0.154$\\
HF vs IO        & CT-Fit    & $0.216$   & $0.186$\\
                & CT-Wide   & $0.294$   & $0.260$\\
\hline
                & CT-Slim   & $0.154$   & $0.148$\\
HF              & CT-Fit    & $0.192$   & $0.168$\\
                & CT-Wide   & $0.261$   & $0.219$\\
\hline
                & CT-Slim   & $0.143$   & $0.140$\\
IO              & CT-Fit    & $0.176$   & $0.163$\\
                & CT-Wide   & $0.251$   & $0.206$\\
\hline
\end{tabular}
\label{tab_similar}
\end{center}
\end{table*}

To investigate the inter-head diversity of our proposed network, we followed a similar procedure of \cite{yun2021analyzing}. We employed the Centered Kernel Alignment (CKA)~\cite{pmlr-v97-kornblith19a} to find the representational similarities $CKA_{(HR)}$ between the heads within each variant. The CKA measures the similarity between $0$ to $1$, with $1$ being the most similar. For an individual validation set with $X$ samples, the representations of every head $h$ were permuted axis-wise and reshaped as $\mathcal{R}^{X.P.T\times D}$ to compute the per-pair unbiased linear CKAs. Finally, samples of all inter-head $CKA_{(HR)}$ arising from every validation set were concatenated. Figure~\ref{fig:8} plots distribution for the samples of $CKA_{(HR)}$ using Kernel Density Estimations (KDEs) for 6-category and 72-exemplar classifications (remaining plots deferred to section S4 of the supplementary material). Further, Table~\ref{tab_similar} expresses the means of the $CKA_{(HR)}$ for different variants and tasks. We notice that except for the CT-Wide for 6-category and 72-exemplar classifications, all other $CKA_{(HR)}$ concentrated toward the $0$'s end with mean similarity values being less than $0.3$, indicating the presence of an implicit diversity amongst different heads.

Interestingly, similarity values are monotonically increasing for the wider variants, consisting of more heads $H$ and larger per-head projection dimension $D$. This behavior is in contrast to the inferences from \cite{yun2021analyzing}. However, this diverging behavior may result from experiments conducted on entirely different tasks and dataset types. Additionally, we highlight the experiments from \cite{pmlr-v97-kornblith19a}, which shows that different wider networks converge to more similar representations. From this, we hypothesize that different heads may diversify their learned representations, but since these representations learn from the linear projections ($q_P, k_P, v_P$) of the initial channels $C$, the increase in $H$ or $D$ may result in significant overlap of per-head representation.


\begin{figure*}[t]
	\centering
	\begin{subfigure}[h]{0.49\textwidth}
		\centering
		\includegraphics[width=\textwidth]{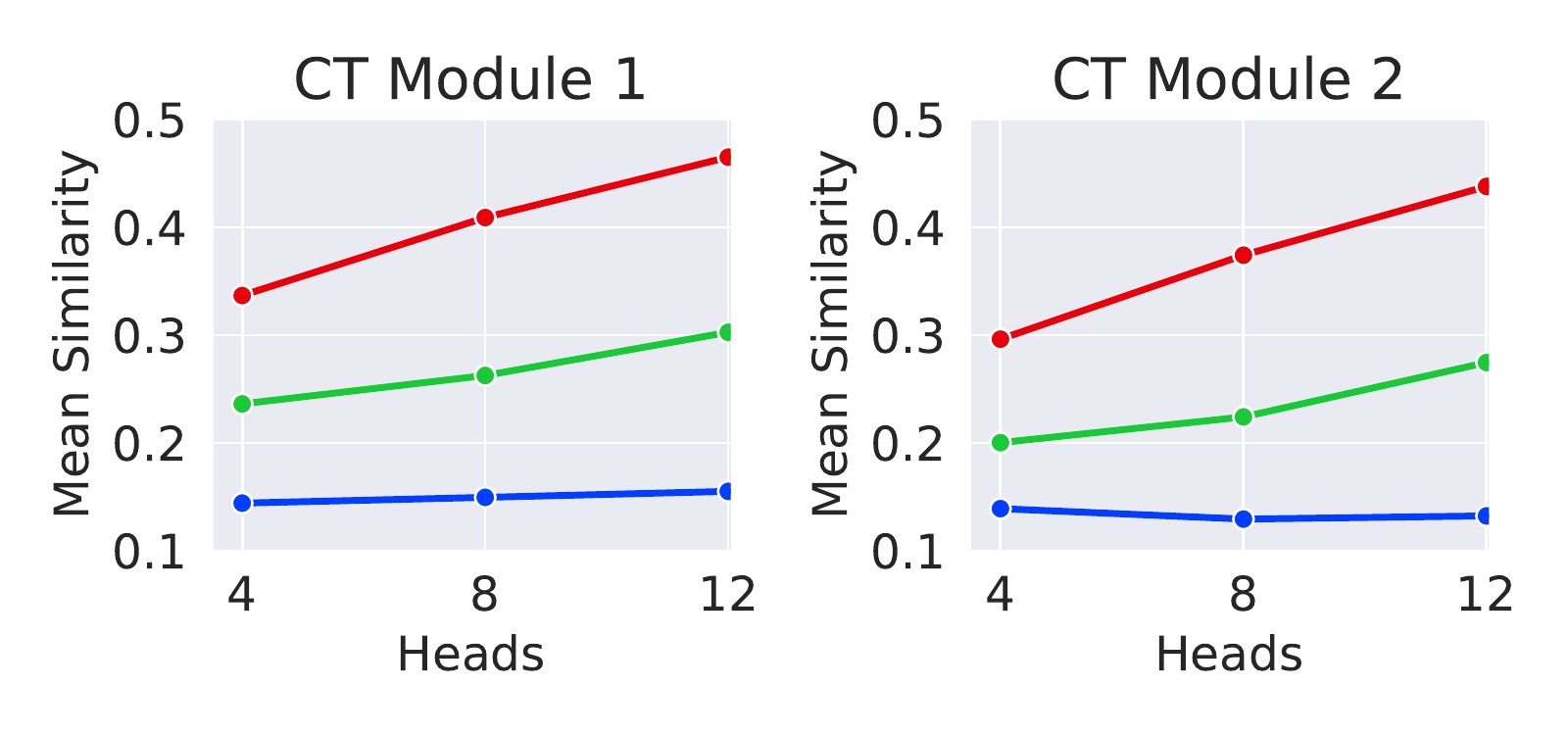}
		\caption{for 6-Category classification}
		\label{fig:9a}
	\end{subfigure}
	\begin{subfigure}[h]{0.49\textwidth}
		\centering
		\includegraphics[width=\textwidth]{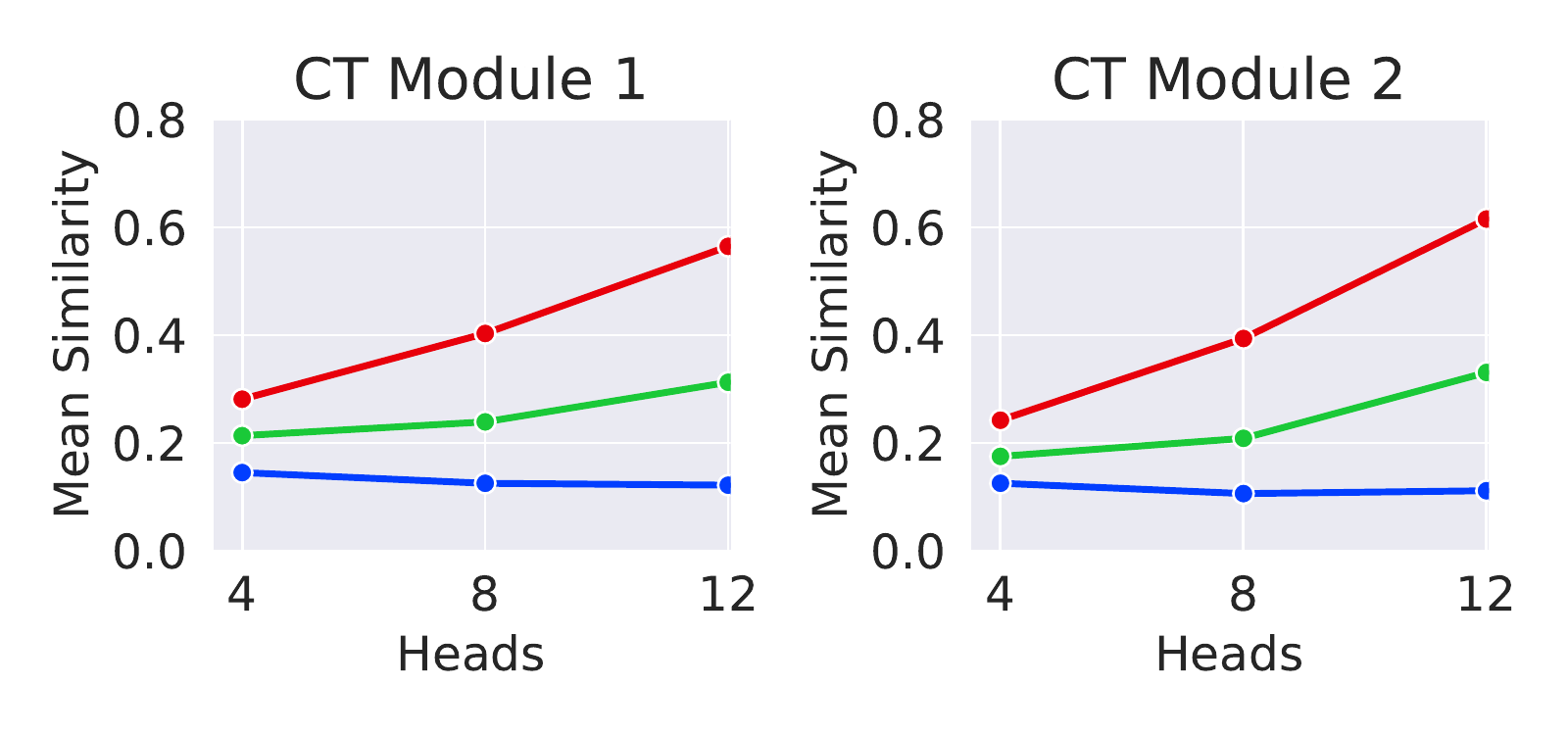}
		\caption{for 72-Exemplar classification}
		\label{fig:9b}
	\end{subfigure}
	\centering
	\begin{subfigure}[h]{0.49\textwidth}
		\centering
		\includegraphics[width=\textwidth]{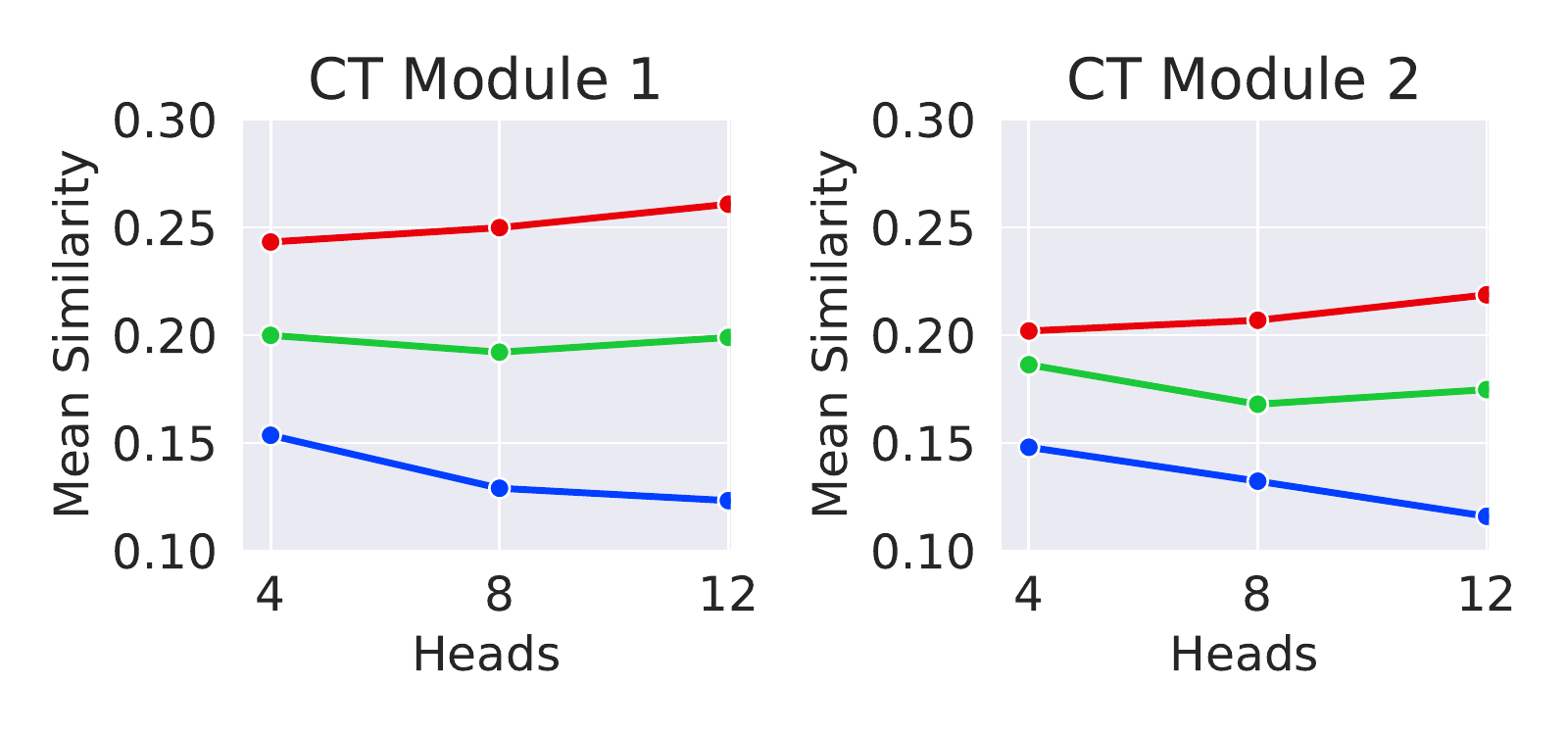}
		\caption{For HF classification}
		\label{fig:9c}
	\end{subfigure}
	\begin{subfigure}[h]{0.49\textwidth}
		\centering
		\includegraphics[width=\textwidth]{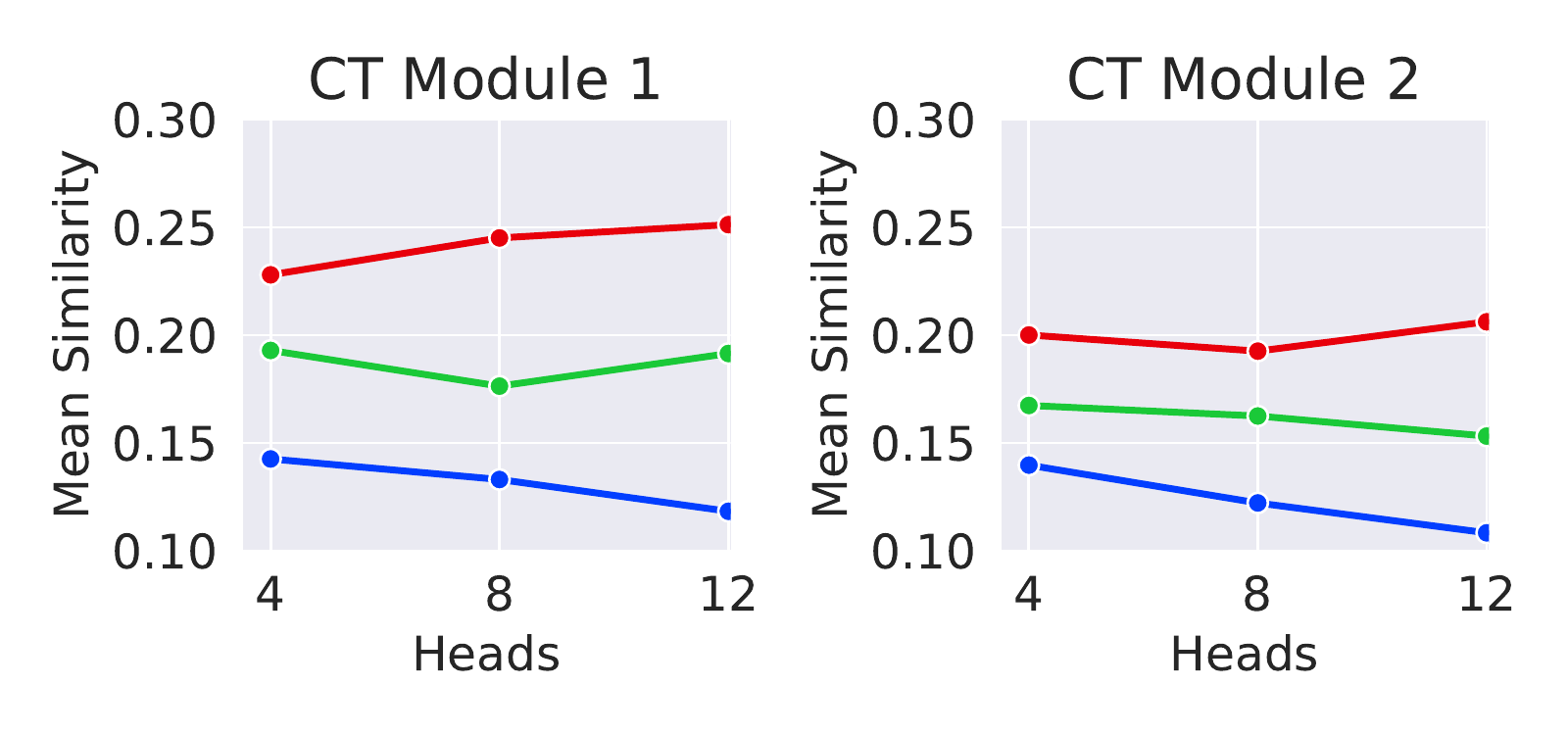}
		\caption{for IO classification}
		\label{fig:9d}
	\end{subfigure}
	\begin{subfigure}[h]{0.6125\textwidth}
		\centering
		\includegraphics[width=\textwidth]{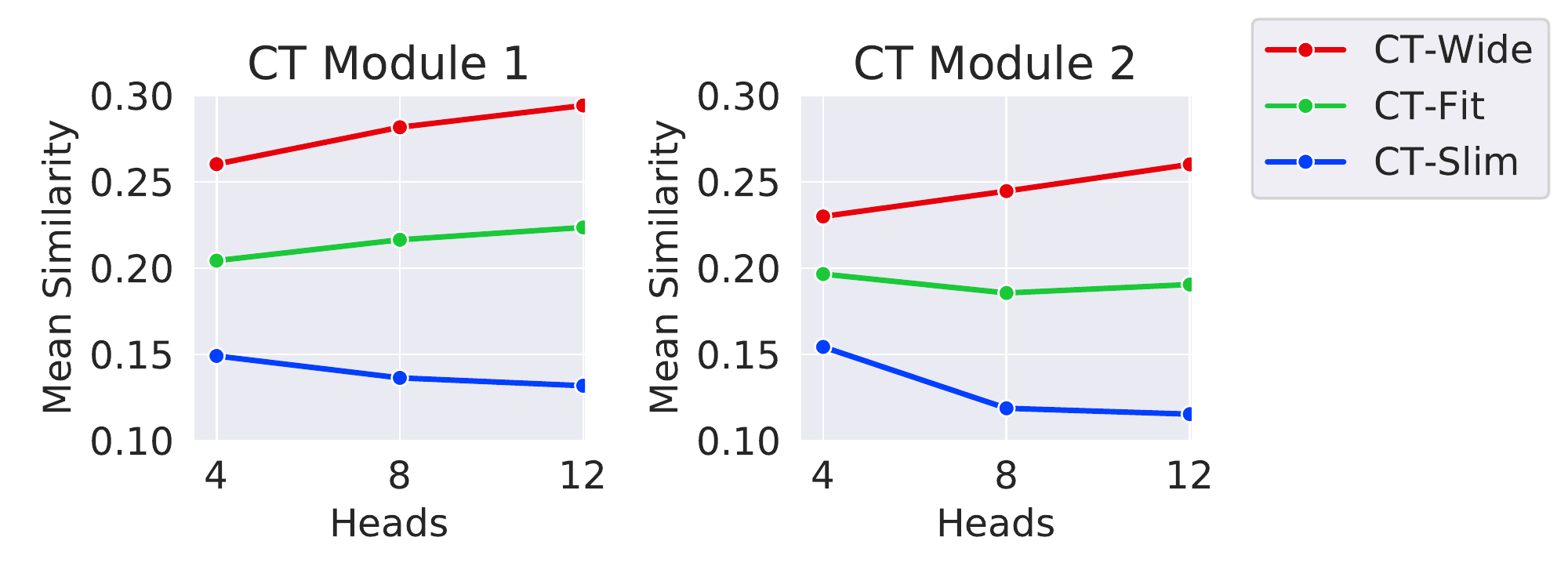}
		\caption{for HF vs IO classification}
		\label{fig:9e}
	\end{subfigure}
	\caption{Mean $CKA_{(HR)}$ of each variants for different number of heads.}
	\label{fig:9}
\end{figure*}

To validate our hypothesis, we conducted further experiments by changing the $H$ and multi-head input channel $C$ (depended on $H$) while preserving the other parameters (including $D$) and hyper-parameters. As perceptible from their mean similarities in Figure~\ref{fig:9}, increasing the $H$ for the CT-Slim decreases the representational similarity (exception being CT Module 1 for 6-category classification). In contrast, the CT-Wide behaves the opposite, except for CT-Module 2 in IO classification. The behavior of CT-Fit, having width in-between the both mentioned above, is rather intricate, and depending on $H$, the similarity could either be decreasing or increasing. Together, the behaviors of the variants suggest that both $H$ and $D$ are vital contributors in determining the inter-head diversity, and beyond a critical channel space, the inter-head diversity may decline.

\section{Discussion}

Our experiments revealed that the improvements led by the different variants of the proposed architecture are more apparent for exemplar-level classifications. Considering the stochastic nature of the training process and the fine-tuning of the hyper-parameters, the minor differences in accuracies between the CT-Fit and CT-Wide, observed for category-level classifications, hold negligible significance. Instead, CT-Wide implores both comprehensive and microscopic level analysis for category-level classification in the future. From the results of our experiments, we hypothesize that the influence of heads and projections is more extensive for exemplar-level classifications.

Our analysis found the nature of inter-head diversity to be piquing. Deep learning models require a lot of fine-tuning for both parameters and hyper-parameters. Although our analysis characterizes the inter-head diversity to the parameters, our hypothesis must be reinforced by several extensive and independent experiments to make a firm inference. Concurrently, the high $CKA_{(HR)}$ of the CT-Wide for 6-category and 72-exemplar classifications also require additional probing.

One limitation of the proposed network is the quadratic computational complexity of the MHA. This is because the self-attention heads rely on the interactions among all the pairs of patches, as explained in section \ref{subsubsection:mha}. Thus, for given $P$ patches, the computational complexity results in $\mathcal{O}(P^2)$. For this reason, if we are to form a mesh of higher resolution, for instance, doubling the height and width of the mesh (resulting in $64\times64$) and keeping the kernel size $k_m$ in $M^{[LFE]}$ constant, results in $P=15\times15=225$ compared to the proposed $P=49$. As a result, the computation increases $\sim$21 times. Similarly, if we are to decrease the kernel size $k_m$, the same problem arises. As highlighted in section \ref{subsection:convtransformer}, the problem is also persistent for inter-channel interactions for a high-density recording device instead of the regions. In the future, if heuristics are introduced for calculating the self-attentions, it should increase the flexibility and scalability of the architecture.

In the future, we would also like to extend our experiments to other cognitive classifications for data acquired through devices with different electrode densities. These experiments would allow for a holistic performance evaluation while analyzing the appropriate patch size and interpolation. Finally, this work could also be extended to other brain imaging modalities like fMRI, where patches of voxels could act as inputs for the self-attention heads. Given that fMRI data have a better spatial resolution, the proposed architecture could be beneficial if provided with enough training data.

\section{Conclusion}
\label{conclusion}

In summary, this work proposes a novel deep learning architecture for EEG-based visual stimuli classification that combines the transformer and convolutional neural networks in a single module to capture both the inter-region similarities and the inherent temporal properties. The proposed network leverages the benefit of introducing multi-head attention to the spatial domain. Three variants of the proposed architecture were explored to analyze the sensitivity of the model to varying parameters including number of heads and the width of the network. Experimental results demonstrate the efficacy of the proposed model in improving the classification accuracy across five different classification tasks when compared to previous methods. Finally, post experimentation analysis on the inter-head diversity concluded implicit diversity amongst the heads depending on the parameter space involved.

\printbibliography

\newpage

\section*{S1 Training details}

For the training of the models, the learning rate and the batch size were taken to be $1e-4$ and 64, respectively. A multi-step learning rate scheduler was further employed to decrease the learning rate by a multiplicative decay $\gamma$ from the \textbf{$15^{th}$} epoch and after every 5 epochs henceforth. Table~\ref{tab_hparam} shows the hyper-parameters that were further fine-tuned for the training of the models for different tasks.

\begin{table}[h]
\caption{Training hyper-parameters for different tasks with respect to each of the variant}
\begin{center}
\begin{tabular}{p{0.16\textwidth}p{0.13\textwidth}p{0.1\textwidth}p{0.1\textwidth}p{0.1\textwidth}}
\hline
\textbf{Parameters} & \textbf{Task} & \textbf{CT-Slim} & \textbf{CT-Fit} & \textbf{CT-Wide}\\
\hline
                        & 6-Category    & 35 & 35 & 40 \\
                        & 72-Exemplar   & 80 & 35 & 35 \\
Training Epochs         & HF vs IO      & 70 & 70 & 10 \\
                        & HF            & 70 & 70 & 10 \\
                        & IO            & 70 & 70 & 10 \\
\hline
                        & 6-Category    & 0.135 & 0.180 & 0.170 \\
                        & 72-Exemplar   & 0.016 & 0.030 & 0.035 \\
Weight Decay            & HF vs IO      & 0.600 & 0.750 & 0.750 \\
                        & HF            & 0.100 & 0.400 & 0.400 \\
                        & IO            & 0.150 & 0.350 & 0.375 \\
\hline
                        & 6-Category    & 0.5 & 0.5 & 0.6 \\
                        & 72-Exemplar   & 0.7 & 0.7 & 0.7 \\
$\gamma$                & HF vs IO      & 0.6 & 0.6 & 0.6 \\
                        & HF            & 0.6 & 0.6 & 0.6 \\
                        & IO            & 0.6 & 0.6 & 0.6 \\
\hline
\end{tabular}
\label{tab_hparam}
\end{center}
\end{table}

\newpage

\section*{S2 Subject wise accuracies}

\begin{figure*}[h]
	\centering
	\begin{subfigure}[h]{.75\textwidth}
		\centering
		\includegraphics[width=\textwidth]{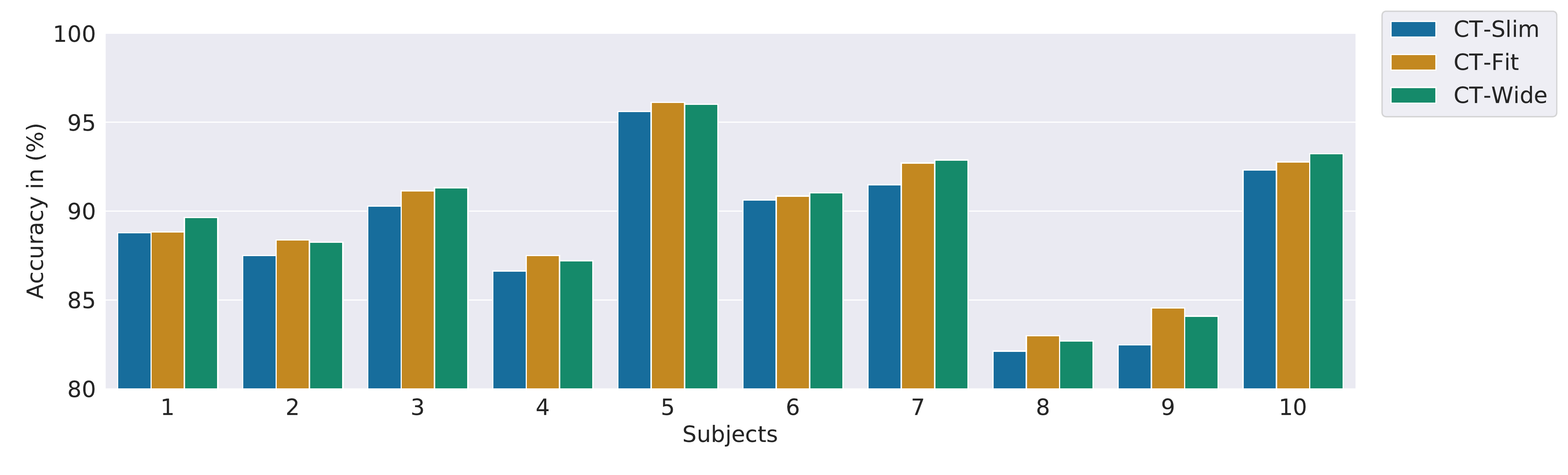}
		\caption{for HF vs IO classification}
		\label{fig:s1a}
	\end{subfigure}
	\begin{subfigure}[h]{.75\textwidth}
		\centering
		\includegraphics[width=\textwidth]{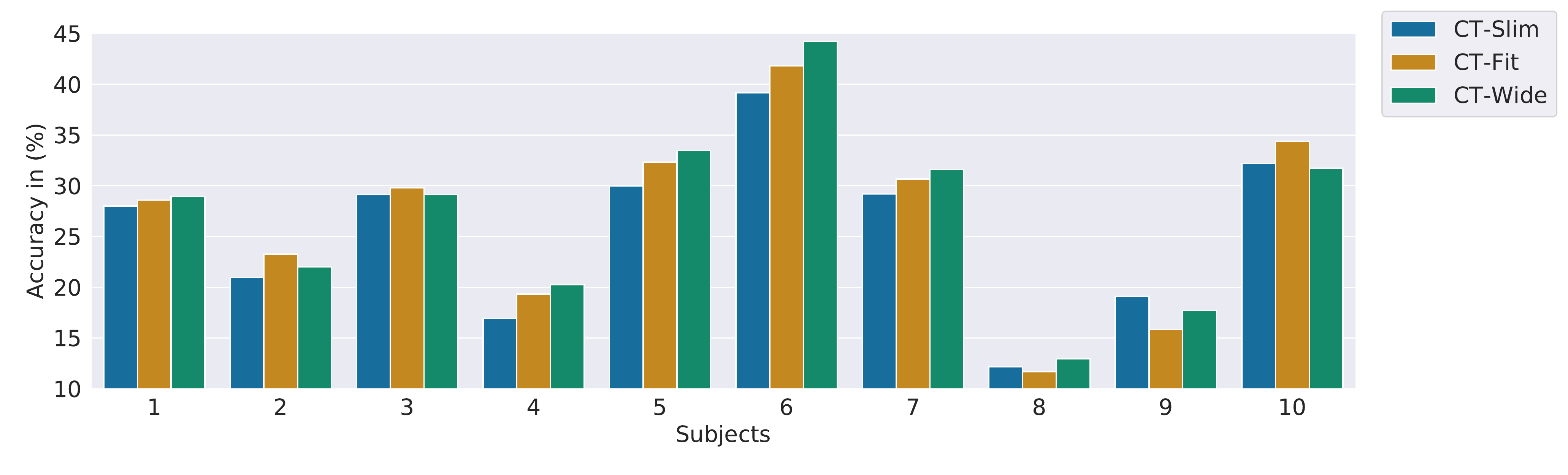}
		\caption{for HF exemplar classification}
		\label{fig:s1b}
	\end{subfigure}
	\begin{subfigure}[h]{.75\textwidth}
		\centering
		\includegraphics[width=\textwidth]{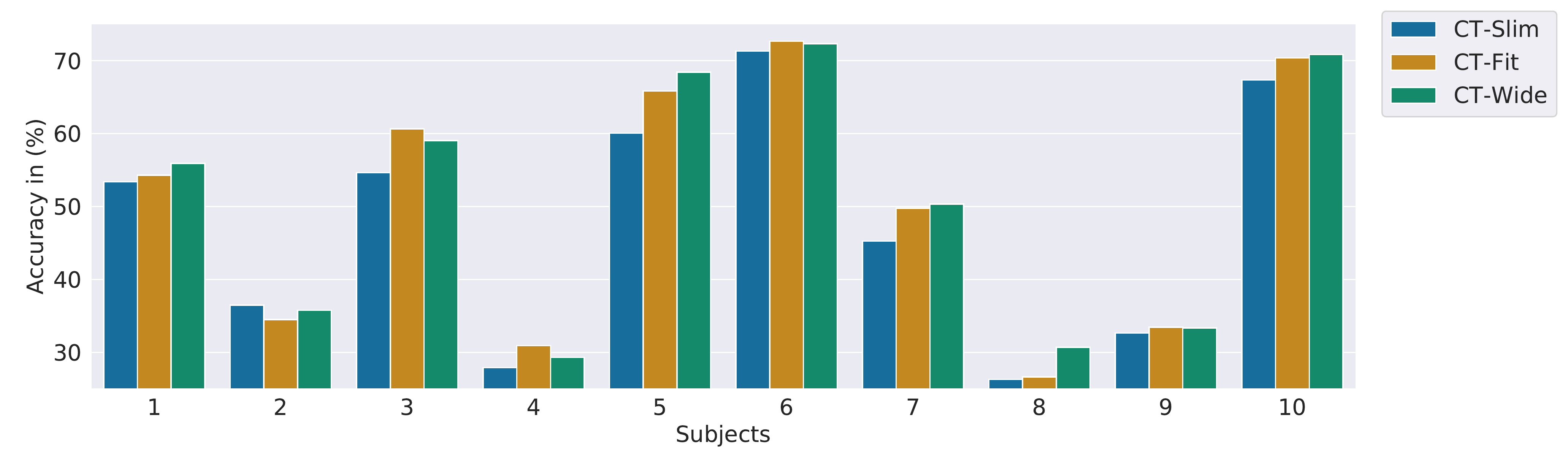}
		\caption{for IO exemplar classification}
		\label{fig:s1b}
	\end{subfigure}
	\caption{Remaining plots for subject wise classification accuracies.}
	\label{fig:s1}
\end{figure*}

\newpage

\section*{S3 Confusion matrices}

\begin{figure*}[h]
	\centering
	\begin{subfigure}[h]{0.4\textwidth}
		\centering
		\includegraphics[width=\textwidth]{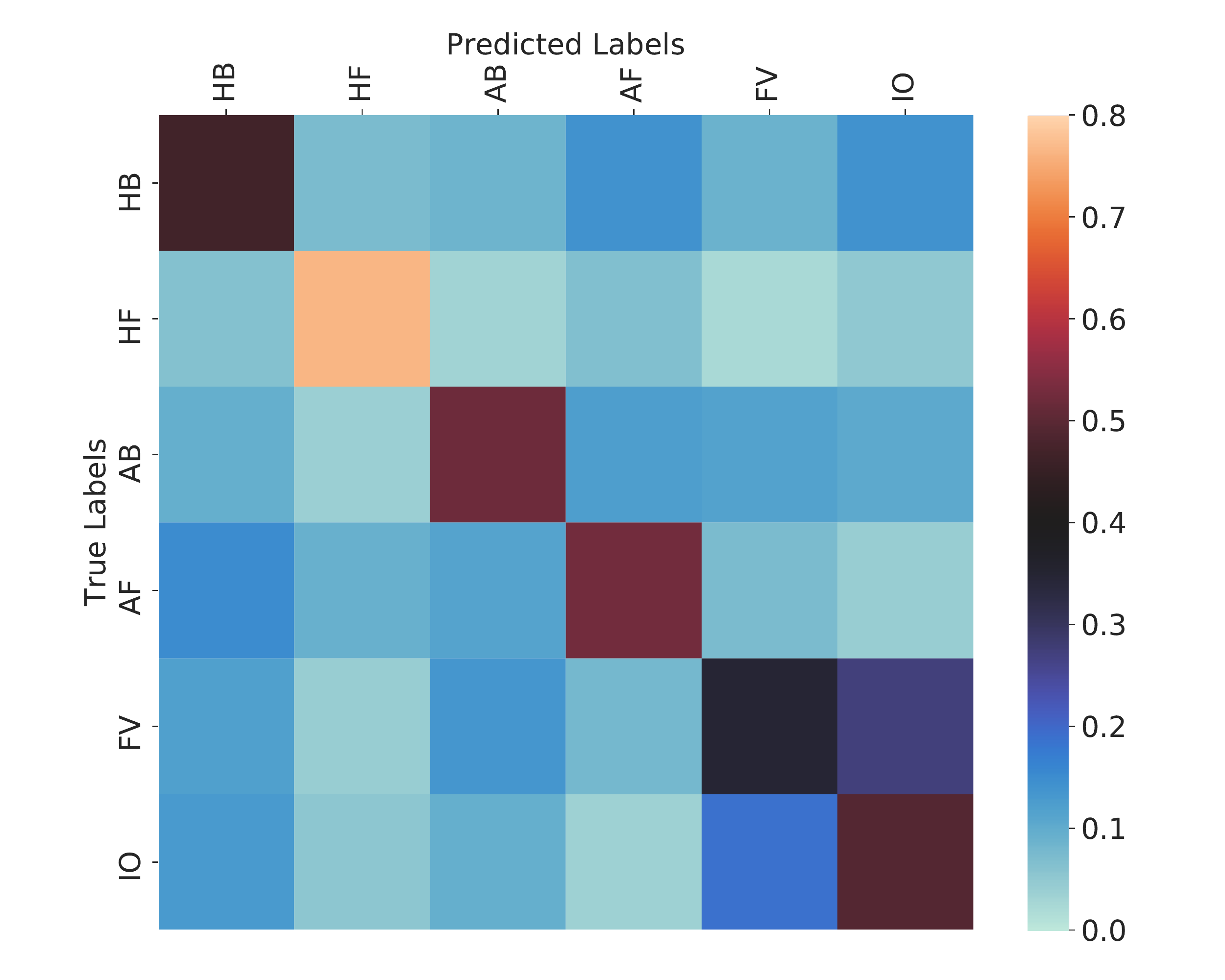}
		\caption{for CT-Slim}
		\label{fig:s2a}
	\end{subfigure}
	\begin{subfigure}[h]{0.4\textwidth}
		\centering
		\includegraphics[width=\textwidth]{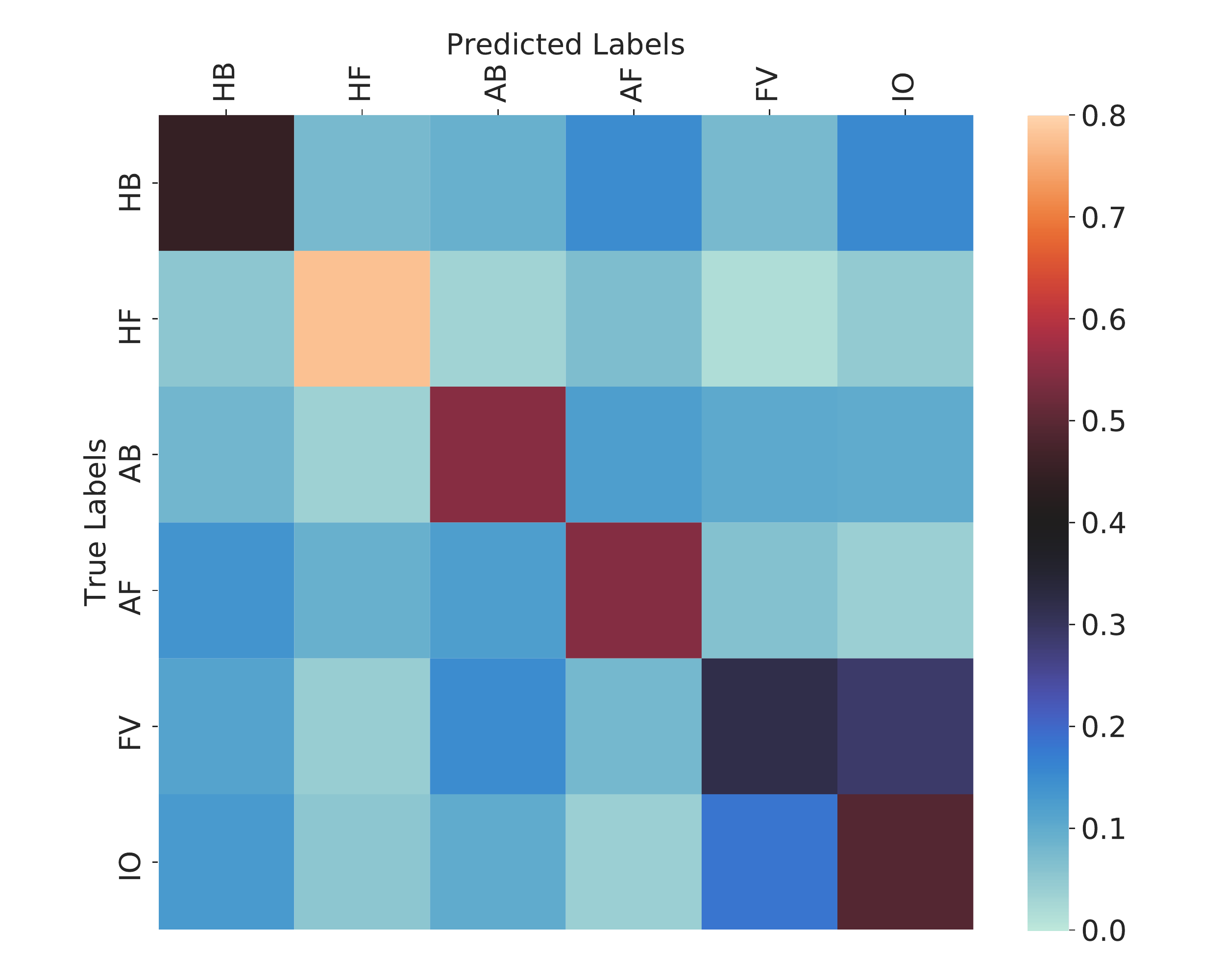}
		\caption{for CT-Fit}
		\label{fig:s2b}
	\end{subfigure}
	\caption{6-category confusion matrices for CT-Slim and CT-Fit.}
	\label{fig:s2}
\end{figure*}

\begin{figure*}[h]
	\centering
	\begin{subfigure}[h]{0.4\textwidth}
		\centering
		\includegraphics[width=\textwidth]{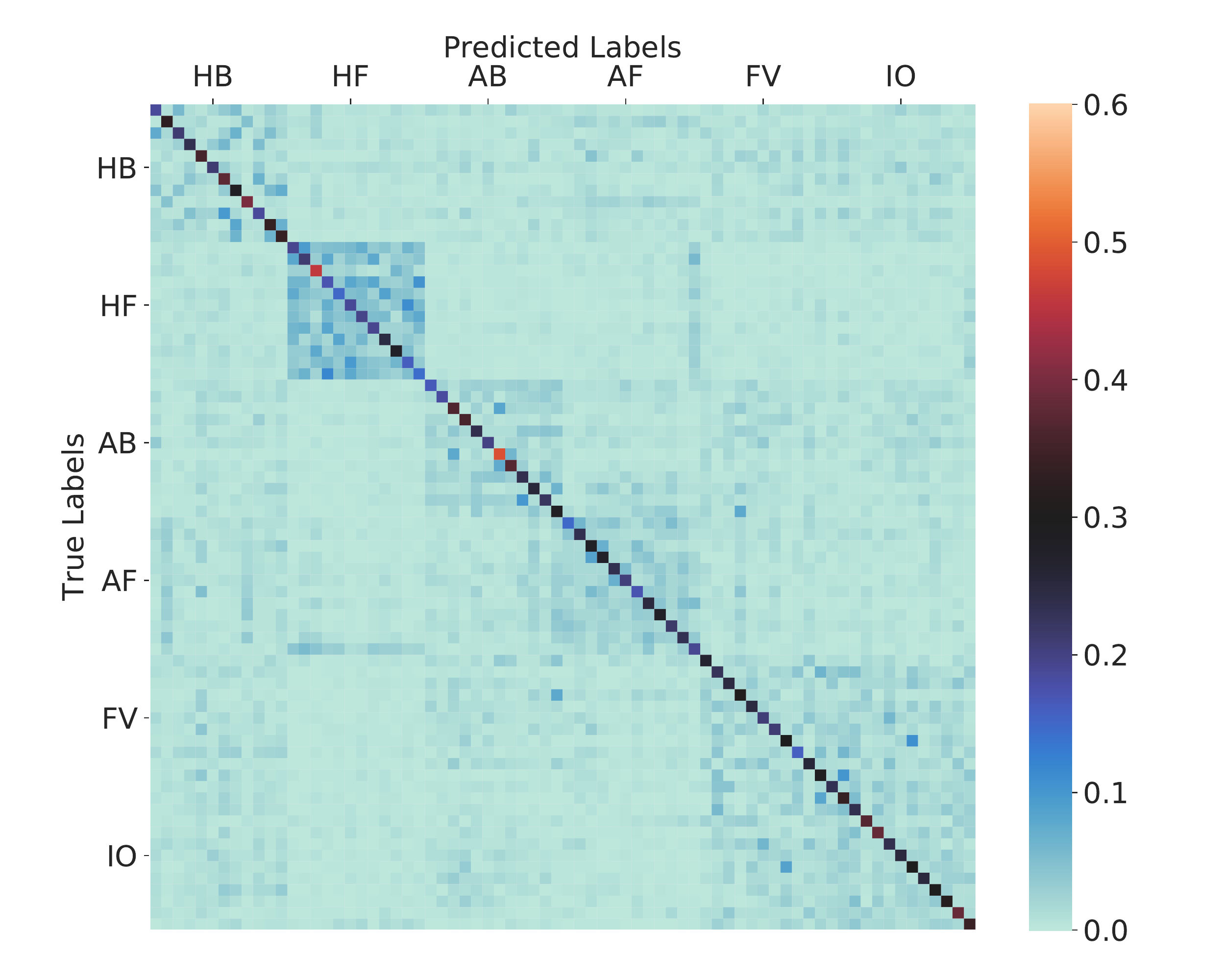}
		\caption{for CT-Slim}
		\label{fig:s_2a}
	\end{subfigure}
	\begin{subfigure}[h]{0.4\textwidth}
		\centering
		\includegraphics[width=\textwidth]{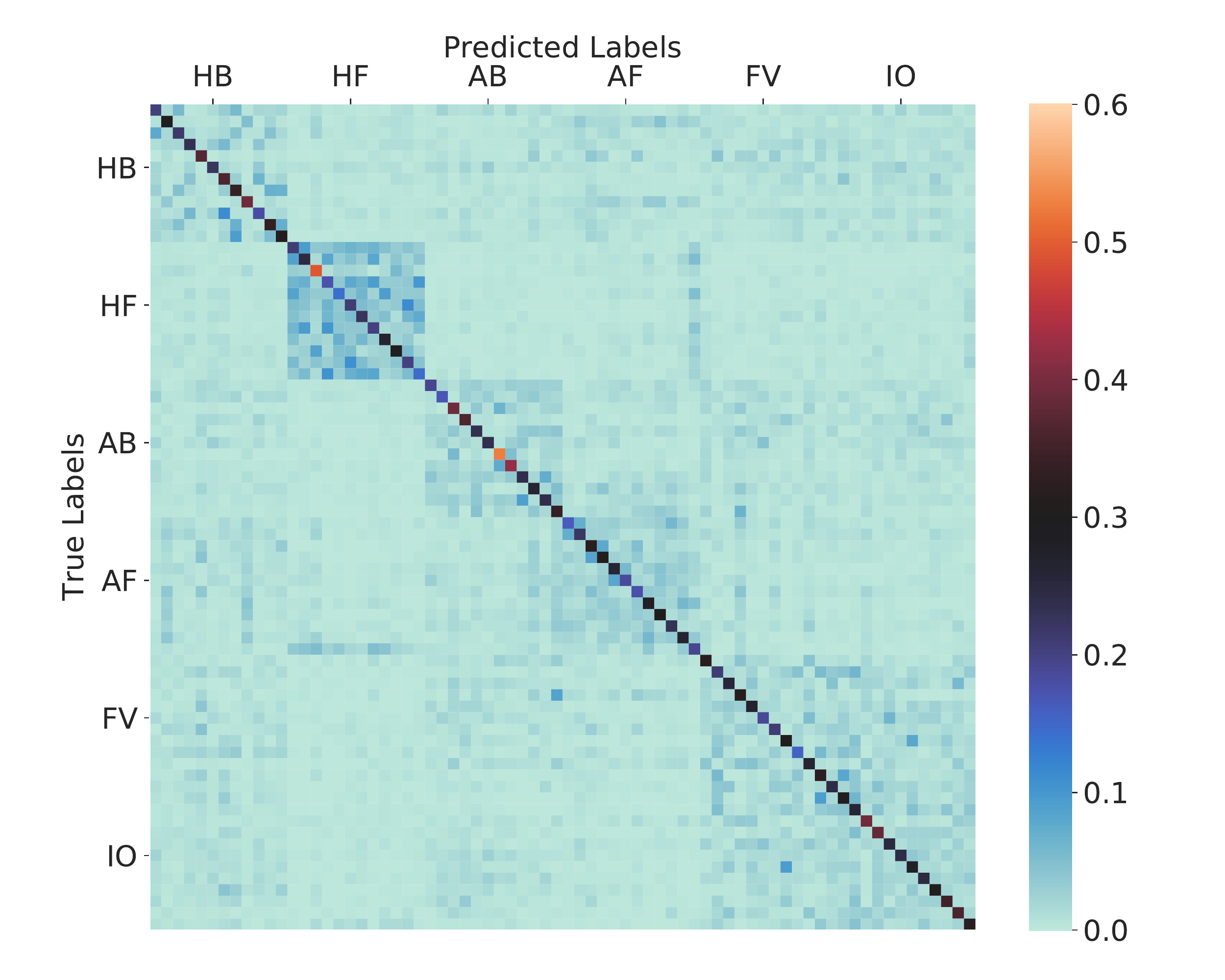}
		\caption{for CT-Fit}
		\label{fig:s_2b}
	\end{subfigure}
	\caption{72-exemplar confusion matrices for CT-Slim and CT-Fit.}
	\label{fig:s_2}
\end{figure*}

\newpage

\section*{S4 Inter-head diversity}

For analysing the per-layer inter-head diversity $CKA_{(HR)}$, the variants \textbf{CT-Slim}, \textbf{CT-Fit} and \textbf{CT-Wide} resulted in 6, 28 and 66 number of inter-head pairs per single validation set, respectively. Thus, for 10 fold cross-validation and for all the 10 subjects, the variants \textbf{CT-Slim}, \textbf{CT-Fit} and \textbf{CT-Wide} have total 600, 2800 and 6600 number of inter-head pairs, respectively. The similarities from these pairs are plotted using Kernel Density Estimations (KDEs).

\begin{figure*}[h]
	\centering
	\begin{subfigure}[h]{0.75\textwidth}
		\centering
		\includegraphics[width=\textwidth]{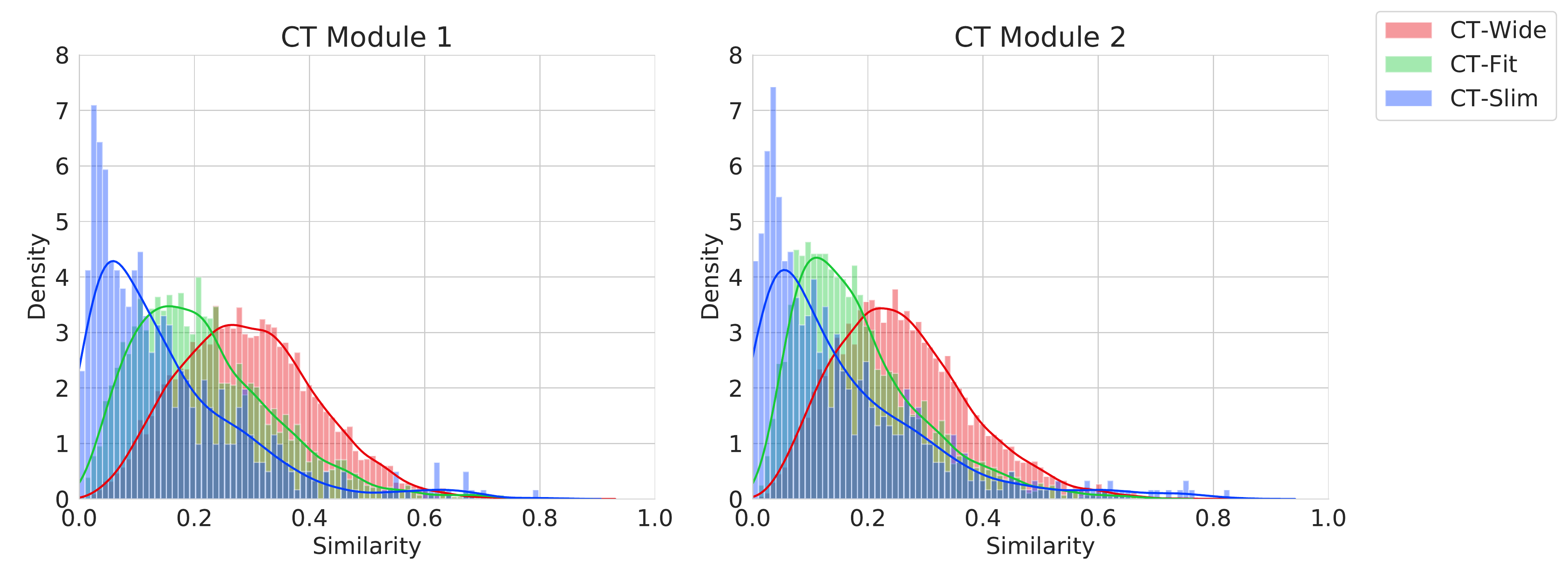}
		\caption{for HF vs IO classification}
		\label{fig:s3a}
	\end{subfigure}
	\begin{subfigure}[h]{0.75\textwidth}
		\centering
		\includegraphics[width=\textwidth]{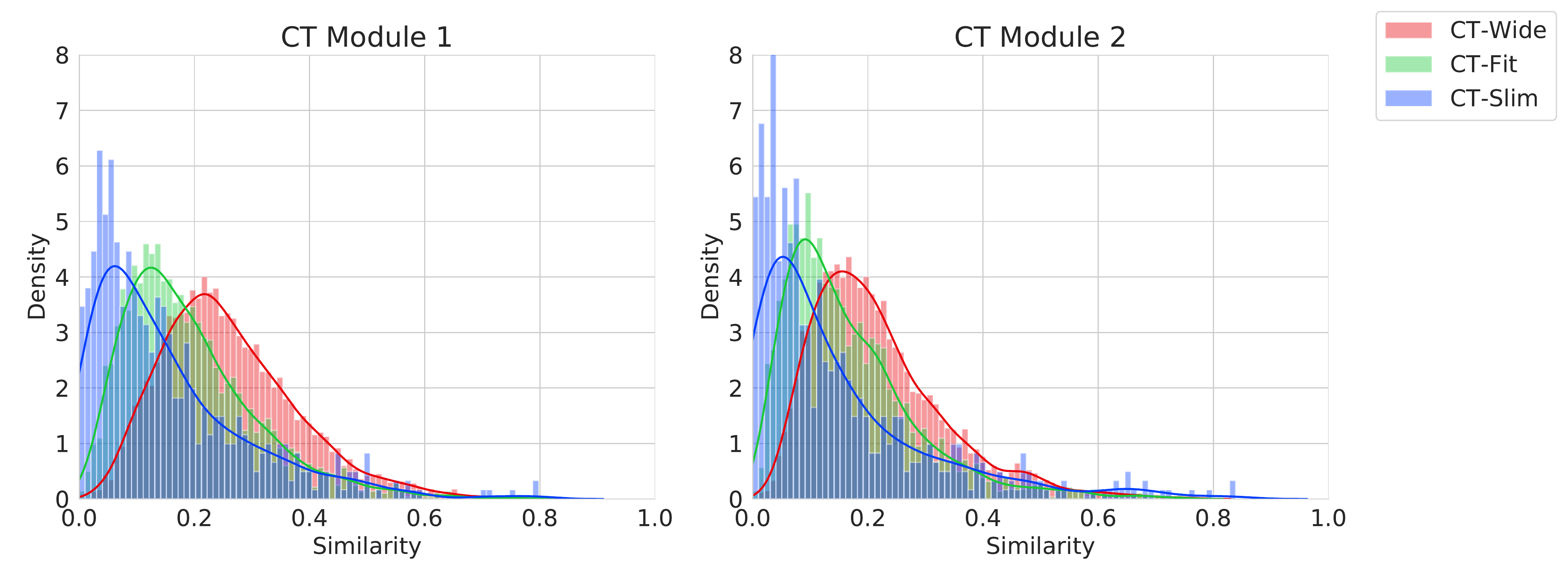}
		\caption{for HF exemplar classification}
		\label{fig:s3b}
	\end{subfigure}
	\begin{subfigure}[h]{0.75\textwidth}
		\centering
		\includegraphics[width=\textwidth]{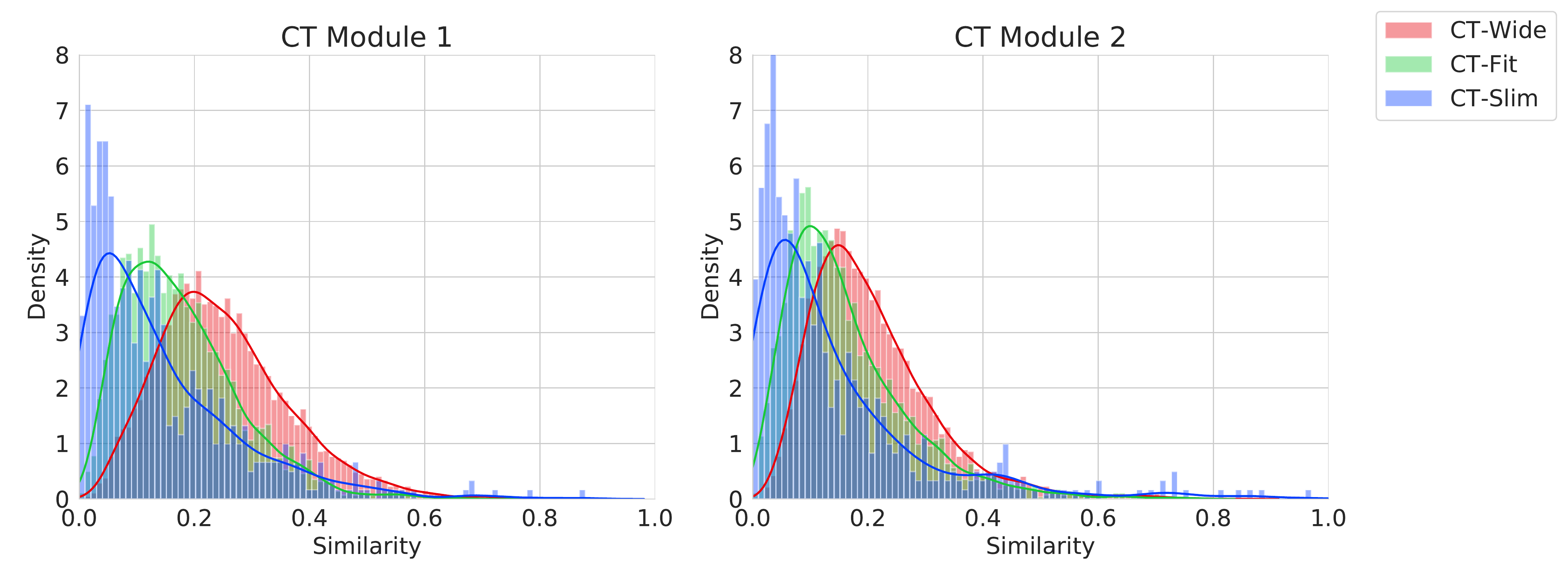}
		\caption{for IO exemplar classification}
		\label{fig:s3c}
	\end{subfigure}
	\caption{$CKA_{(HR)}$ sample distribution using KDEs for the remaining tasks.}
	\label{fig:s3}
\end{figure*}

\end{document}